\documentclass[10pt,twocolumn,letterpaper]{article}

\usepackage{iccv}
\usepackage{times}
\usepackage{epsfig}
\usepackage{graphicx}
\usepackage{amsmath}
\usepackage{amssymb}

\usepackage{color,xcolor}
\usepackage{booktabs}
\usepackage[utf8]{inputenc}
\usepackage{multirow}
\usepackage{makecell}

\usepackage{subcaption}
\usepackage{soul}
\usepackage{rotating}
\usepackage{pdflscape}
\usepackage{afterpage}
\usepackage{flafter}
\usepackage{fancyhdr}

\fancypagestyle{lscape}{% 
    \fancyhf{} % clear all header and footer fields 
    \fancyfoot[LE]{}
    \fancyfoot[LO] {}
    \renewcommand{\headrulewidth}{0pt} 
    \renewcommand{\footrulewidth}{0pt}
}

\newcommand\blfootnote[1]{%
  \begingroup
  \renewcommand\thefootnote{}\footnote{#1}%
  \addtocounter{footnote}{-1}%
  \endgroup
}

\usepackage{amsmath,amsfonts,bm}

\def\eqref#1{equation~\ref{#1}}
\def\1{\bm{1}}

\def\rvu{{\mathbf{i}}}

\def\rvr{{\mathbf{r}}}

\def\rvu{{\mathbf{u}}}

\def\rvx{{\mathbf{x}}}
\def\rvy{{\mathbf{y}}}
\def\rvz{{\mathbf{z}}}

\DeclareMathAlphabet{\mathsfit}{\encodingdefault}{\sfdefault}{m}{sl}
\SetMathAlphabet{\mathsfit}{bold}{\encodingdefault}{\sfdefault}{bx}{n}

\newcommand{\E}{\mathbb{E}}

\newcommand{\xo}{\mathbf{x_0}}
\newcommand{\xone}{\mathbf{x_1}}

\newcommand{\xt}{\mathbf{x_t}}
\newcommand{\xT}{\mathbf{x_T}}

\newcommand{\xpast}{\mathbf{x_{t-1}}}

\newcommand{\rzero}{\mathbf{r_0}}

\newcommand{\bat}{\Bar{\alpha}_{t}}
\newcommand{\batminus}{\Bar{\alpha}_{t-1}}

\newcommand{\zero}{\boldsymbol{0}}

\newcommand{\e}{\boldsymbol{\epsilon}}

\newcommand{\mutheta}{\boldsymbol{\mu}_\theta}
\newcommand{\tmut}{\boldsymbol{\tilde{\mu}}_t}

\newcommand{\N}{\mathcal{N}}

\newcommand{\target}{\rvx}
\newcommand{\initialrecon}{\mathbf{\tilde{x}}}

\newcommand{\recon}{\mathbf{\hat{x}}}

\newcommand{\residual}{\rvr}

\newcommand{\quantlatent}{\rvy}

\newcommand{\ratetradeoff}{\lambda_{\text{rate}}}

\usepackage[pagebackref=true,breaklinks=true,letterpaper=true,colorlinks,bookmarks=false]{hyperref}
\usepackage[capitalise]{cleveref}

\title{Neural Image Compression with a Diffusion-based Decoder}

\begin{document}

%%%%%%%%%%%%%%%%%%%%%%%%%
\title{A Residual Diffusion Model for High Perceptual Quality Codec Augmentation}

\iccvfinalcopy % *** Uncomment this line for the final submissionk

% Order and exact list of authors TBD
\author{
Noor Fathima Ghouse*,  
Jens Petersen*, 
Auke Wiggers*, 
Tianlin Xu\textsuperscript{$\ddag$},
Guillaume Sauti\`{e}re \\
{
  \small \texttt{\{noor, jpeterse, auke, gsautie\}@qti.qualcomm.com; tianlin.xu1@gmail.com}
}
}

\maketitle

\begin{abstract}
Diffusion probabilistic models have recently achieved remarkable success in generating high quality image and video data.
In this work, we build on this class of generative models and introduce a method for lossy compression of high resolution images.
The resulting codec, which we call \emph{DIffuson-based Residual Augmentation Codec (DIRAC)}, 
is the first neural codec to allow smooth traversal of the rate-distortion-perception tradeoff at test time, 
while obtaining competitive performance with GAN-based methods in perceptual quality.
Furthermore, while sampling from diffusion probabilistic models is notoriously expensive, we show that 
in the compression setting the number of steps can be drastically reduced.
\end{abstract}

%%%%%%%%%%%%%
\blfootnote{
    Preprint. Qualcomm AI Research is an initiative of Qualcomm Technologies, Inc. \textsuperscript{$\ddag$}: Work completed during an internship at Qualcomm AI Research. \textsuperscript{*}: Equal contribution. 
}
\section{Introduction}

\begin{figure}[t]
    \centering
    \includegraphics[width=\linewidth]{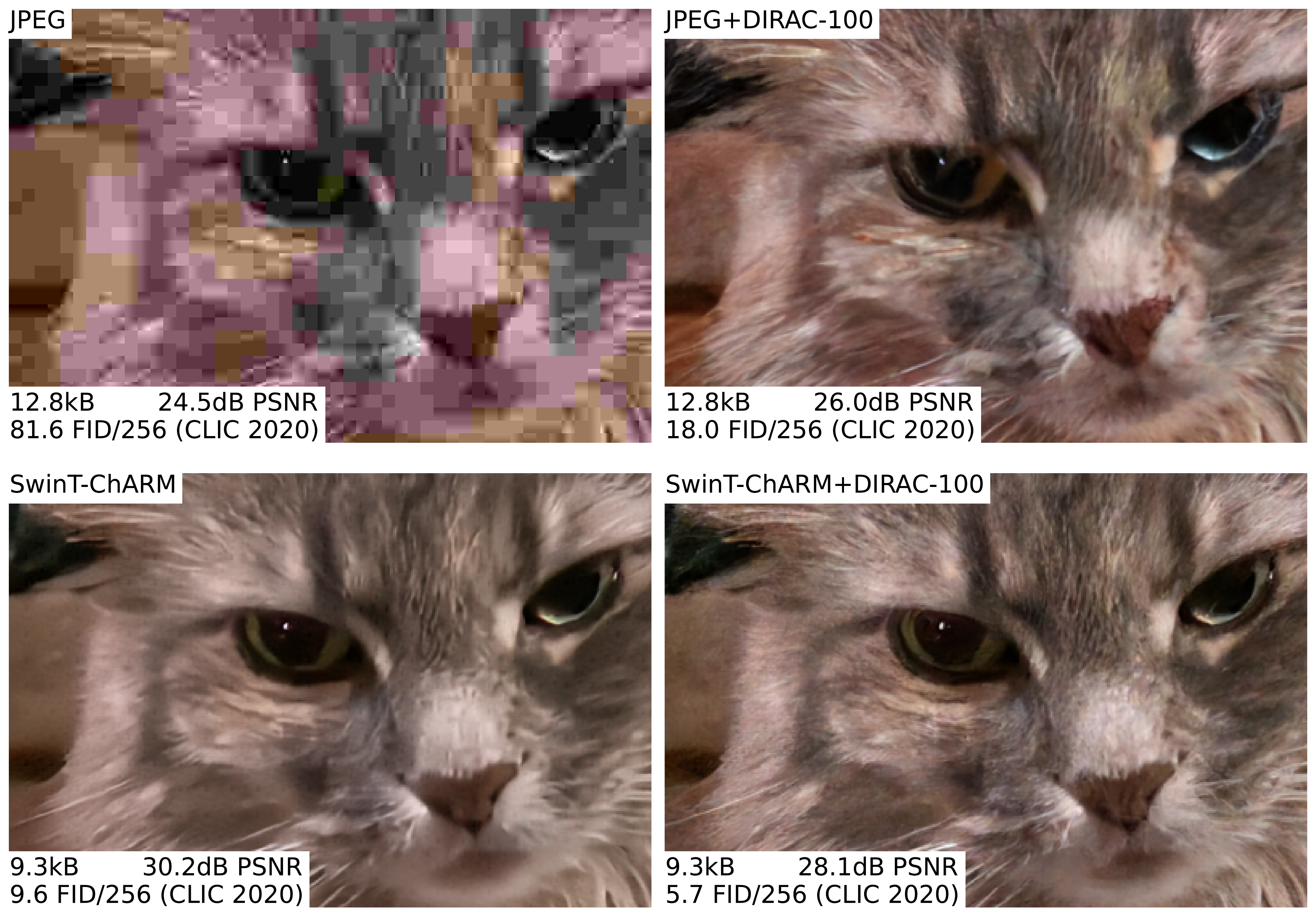}
    \caption{Base reconstruction (left) and the DIRAC-enhanced version (right). 
    Our model combines a base codec with a receiver-side enhancement model, and can smoothly interpolate between near-state-of-the-art fidelity (PSNR, higher better) and near-state-of-the-art perceptual quality (FID, lower better). 
    For JPEG ($QF=5$) specifically, we achieve a drastic improvement in perceptual quality without loss in PSNR. 
    Best viewed digitally, PSNR measured on the shown example, FID/256 measured on the full CLIC 2020 test dataset.}
    \label{fig:hero_figure_example}
\end{figure}

Denoising diffusion probabilistic models (DDPMs)~\cite{sohl2015deep} have recently shown incredible performance in the generation of high-resolution images with high perceptual quality. 
For example, they have powered large text-to-image models such as DALL-E 2~\cite{ramesh2022dalle2} and Imagen~\cite{saharia2022imagen}, which are capable of producing realistic high-resolution images based on arbitrary text prompts. 
Likewise, diffusion models have demonstrated impressive results on image-to-image tasks such as super-resolution \cite{saharia2021image,ho2021cascaded}, deblurring \cite{whang2022deblurring} or inpainting \cite{saharia2022palette}, in many cases outperforming generative adversarial networks (GANs) \cite{dhariwal2021diffusion}. 
Our goal in this work is to leverage these capabilities in the context of learned compression.

Neural codecs, which learn to compress from example data, are typically trained to minimize distortion between an input and a reconstruction, as well as the bitrate used to transmit the data \cite{theis2017lossy}.
However, optimizing for rate-distortion may result in blurry reconstructions. 
A recent class of generative models focuses instead on improving perceptual quality of reconstructions, either with end-to-end trained neural codecs \cite{agustsson2019extreme,mentzer2020hific}---we refer to such techniques as \emph{generative compression}---or by using a receiver-side \emph{perceptual enhancement} model \cite{kawar2022denoising} to augment the output of standard codecs.
Either approach will usually come at a cost in fidelity, as there is a fundamental tradeoff between fidelity and perceptual quality \cite{blau2019rethinking}.
Finding a good operating point for this tradeoff is not trivial and likely application-dependent.
Ideally, one would like to be able to select this operating point at test time.
But while adaptive rate control is commonly used, few neural image codecs allow trading off distortion and perceptual quality dynamically \cite{iwai2021fidelity, agustsson2022multi}.

In this work, we present a method that allows users to navigate the full rate-distortion-perception tradeoff at test time with a single model. 
Our approach, called \emph{DIffusion Residual Augmentation Codec (DIRAC)}, uses a base codec to produce an initial reconstruction with minimal distortion, and then improves its perceptual quality using a denoising diffusion probabilistic model that predicts residuals, see \cref{fig:hero_figure_example} for an example.
The intermediate samples correspond to a smooth traversal between high fidelity and high perceptual quality, so that sampling can be stopped when a desired tradeoff is reached. 
Recent work \cite{yang2022lossy, pan2022extreme} already demonstrates that a diffusion-based image codec is feasible in practice, but we show that different design choices allow us to outperform their models by a large margin, and enable faster sampling as well as selection of the distortion-perception operating point. 
Our contributions are:

\begin{itemize}
    \item We demonstrate a practical and flexible diffusion-based model that can be combined with any image codec to achieve high perceptual quality compression. 
    Paired with a neural base codec, it can interpolate between high fidelity and high perceptual quality while being competitive with the state of the art in both.
    \item Our model can be used as a drop-in enhancement model for traditional codecs, where we achieve strong perceptual quality improvements. 
    For JPEG specifically, we improve FID/256 by up to $78\%$ without loss in PSNR.
    \item We present techniques that make the diffusion sampling procedure more efficient: we show that in our setting we need no more than 20 sampling steps, and we introduce \emph{rate-dependent thresholding}, which improves performance for multi-rate base codecs. 
\end{itemize}

\begin{figure*}[t]
    \centering
    \includegraphics[width=0.9\textwidth]{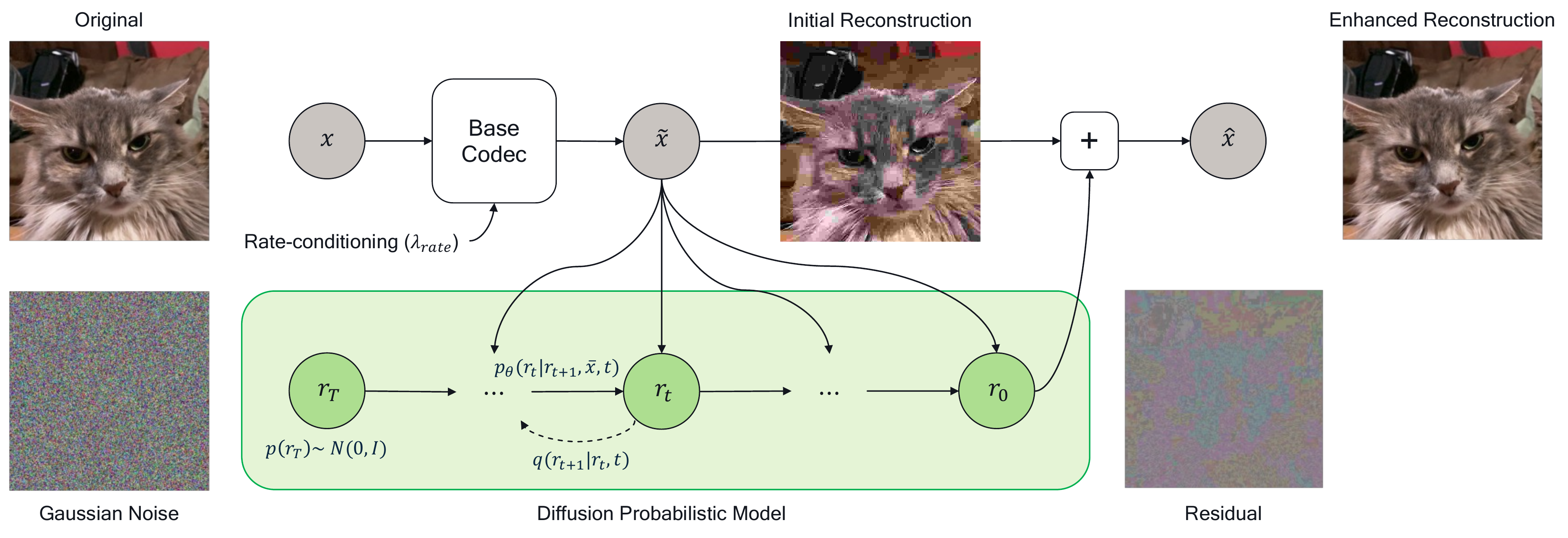}
    \caption{Overview of our architecture. Given an input image $\target$ and target rate factor $\lambda_{rate}$, we obtain a base codec reconstruction $\initialrecon$. Our DDPM is conditioned on $\initialrecon$ and learns to model a reverse diffusion process that generates residuals $r_0$ from sampled gaussian noise latents $r_T$. The enhanced reconstruction $\recon$ is then obtained by adding the predicted residual to $\initialrecon$}
    \label{fig:method:architecture}
\end{figure*}

\section{Related work} 
\label{sec:related_work}

%%%%%%%%%%%%%%%%%%%%%%%%%%%%%%%%%%%%%%%%%%
\paragraph{Neural data compression}

Neural network based codecs are systems that learn to compress data from examples.
These codecs have seen major advances in both the image \cite{minnen2017spatially, rippel2017realtime, balle2018variational, minnen2018joint} and video domain \cite{wu2018video, lu2019dvc, rippel2019lvc, habibian2019video, agustsson2020ssf, hu2021fvc, zhihao2022c2f}.
Most modern neural codecs are variations of \emph{compressive autoencoders} \cite{theis2017lossy}, which transmit data $\target$ using an autoencoder-like architecture, resulting in a reconstruction $\initialrecon$.
These systems are typically optimized using a rate-distortion loss, i.e. a combination of distortion and a rate loss that are balanced with a rate-parameter $\ratetradeoff$.

Recent work identifies the importance of a third objective: \emph{perceptual quality}, ideally meaning that reconstructions look like real data according to human observers.
Blau and Michali \cite{blau2019rethinking} formalize perceptual quality as a distance between the image distribution $p(\target)$ and the distribution of reconstructions $p(\recon)$, 
and show that rate, distortion and perception are in a triple tradeoff.
To optimize for perceptual quality, a common choice is to train the decoder as a (conditional) generative adversarial network by adding a GAN loss term to the rate-distortion loss and training a discriminator \cite{agustsson2019extreme, mentzer2020hific, yang2021perceptual, agustsson2022multi, muckley2023improving}.

It is impractical to train and deploy one codec per rate-distortion-perception operating point.
A common choice is therefore to condition the network on the bitrate tradeoff parameter $\lambda_\text{rate}$, and vary this parameter during training \cite{wu2020gan, song2021variable, rippel2021elfvc}. Recent GAN-based works use similar techniques to trade off fidelity and perceptual quality, either using control parameters, or by masking the transmitted latent \cite{wu2020gan, iwai2021fidelity, agustsson2022multi}. 
In particular, the work by Agustsson \etal \cite{agustsson2022multi} uses receiver side conditioning to trade-off distortion and realism at a particular bitrate.
However, a codec that can navigate all axes of the rate-distortion-perception tradeoff simultaneously using simple control parameters does not exist yet. 

%%%%%%%%%%%%%%%%%%%%%%%%%%%%%%%%%%%%%%%%%%%%%%%%
\paragraph{Diffusion probabilistic models}
\label{sub:ddpm}

Denoising diffusion probabilistic models (DDPMs) \cite{sohl2015deep,ho2020denoising} are latent variable models in which the latents $\xone, ..., \xT$  are defined as a $T$-step Markov chain with Gaussian transitions. 
Through this Markov chain, the \emph{forward process} gradually corrupts the original data $\xo$.
The key idea of DDPMs is that, if the forward process permits efficient sampling of any variable in the chain, we can construct a generative model by learning to reverse the forward process. 
To generate a sample, the reverse model is applied iteratively.
For a thorough description of DDPMs, we refer the reader to
the appendix or Sohl-Dickstein \etal \cite{sohl2015deep}.

DDPMs have seen success in various application areas, including image and video generation \cite{ho2021cascaded, ho2022video, Yang_Srivastava_Mandt_2022, ramesh2022dalle2, saharia2022imagen}, representation learning \cite{preechakul2022diffusion, pandey2022diffusevae}, and image-to-image tasks such as super-resolution \cite{saharia2021image} and deblurring \cite{whang2022deblurring}.
Recent work applies this model class in the context of data compression as well.
Hoogeboom \etal \cite{hoogeboom2021autoregressive} show that DDPMs can be used to perform lossless compression. 
In the lossy compression setting, Ho \etal \cite{ho2020denoising} show that if continuous latents could be transmitted, a DDPM enables progressive coding. 
Theis \etal \cite{theis2022lossy} make this approach feasible by using reverse channel coding,
yet it remains impractical for high resolution images due to its high computational cost.

More practical diffusion-based approaches for lossy image compression exist, too.
Yang and Mandt \cite{yang2022lossy} propose a codec where a conditional DDPM takes the role of the decoder, directly producing a reconstruction from a latent variable.
Pan \etal \cite{pan2022extreme} similarly use a pretrained text-conditioned diffusion model as decoder, and let the encoder extract a text embedding.
However, both approaches still require a large number of sampling steps, or encoder-side optimization of the compressed latent.

%%%%%%%%%%%%%%%%%%%%%%%%%%%%%%%%%%%%%%%%%%%%%
\paragraph{Standard codec restoration}

A common approach is to take a standard codec such as JPEG, and enhance or \emph{restore} its reconstructions. 
Until recently, most restoration works focused on improving distortion metrics such as PSNR or SSIM \cite{dong2015compression,kwak2018image,liu2020comprehensive,ehrlich2020quantization, yang2021ntire, yang2022ntire}. 
However, distortion-optimized restoration typically leads to blurry images, as blur get rids of compression artifacts such as blocking and ringing.
Consequently, a recent category of work on \emph{perceptual enhancement} \cite{kawar2022denoising,kawar2022jpeg,saharia2022palette,wang2022perceptual,song2023solving} focuses mainly on realism of the enhanced image.
This is usually measured by perceptual distortion metrics such as LPIPS \cite{zhang2018lpips} or distribution-based metrics like FID \cite{heusel2017fid}.
%at the cost of PSNR. 
Although enhanced images are less faithful to the original than the non-enhanced version, they may be rated as more realistic by human observers. 
In this setting, DDPMs have mostly been applied to JPEG restoration \cite{kawar2022jpeg, saharia2022palette, song2023solving}. 
However, these methods have not shown test-time control of perception-distortion tradeoff, and were only tested for a single standard codec (JPEG) on low-resolution data.

%%%%%%%%%%%%%%%%%%%%%%%%%%%%%%%%%%%%%%%%%%%%%%%%%%%%%%%
\section{Method}

In this work, we introduce DIRAC, a diffusion-based image compression approach for high-resolution images. 
It combines a (potentially learned) base codec with a residual diffusion model that performs iterative enhancement.
This setup is shown in \cref{fig:method:architecture}.
By design, we obtain both a high fidelity initial reconstruction, and a high perceptual quality enhanced reconstruction.
The enhancement is performed on the receiver side and can be stopped at any time, enabling test-time control over the distortion-perception tradeoff.

%%%%%%%%%%%%%%%%%%%%%%
\subsection{Residual diffusion models}
Diffusion-based enhancement is typically achieved by conditioning the reverse model on the image-to-enhance $\initialrecon$, effectively modeling the conditional distribution $p(\xo | \initialrecon)$ \cite{ho2021cascaded, saharia2022palette, song2023solving}. 
Following Whang \etal \cite{whang2022deblurring}, we instead opt to model the distribution of residuals $p(\rzero | \initialrecon)$, where $\rzero = \target - \initialrecon$ and the index is for conceptual diffusion time.
From an information theory perspective modeling residuals is equivalent to modeling images, but residuals follow an approximately Gaussian distribution, which we believe can be easier to model. 
More details on this choice are given in the appendix.

For training, we use the common loss parametrization where the model learns to predict the initial sample instead of the noise that was added to it. Yang and Mandt \cite{yang2022lossy} note that optimizing the perceptual distortion metric LPIPS \cite{zhang2018lpips} contributes to perceptual performance, and we adopt a similar practice here by adding a loss term, so that our final loss becomes:

\begin{align}
  \mathcal{L}(\target, \initialrecon) 
  = \mathop{\mathbb{E}}_{t, \mathbf{r_t}} 
    \left[  w_t||\rzero - \rzero' ||^2  \, + 
      \lambda_\text{LPIPS} \, d_\text{LPIPS}(\target, \initialrecon + \rzero' )
    \right],
  \label{eq:method:rdploss}
\end{align}

\noindent where $\rzero' = g_\theta(\mathbf{r_t}, t)$ is the prediction from our model. $w_t$ is a weighting factor for which the theoretically derived terms become very large for small $t$ (see appendix for the derivation), so we choose to set $w_t=1$ to balance all loss terms evenly, similar to how Ho \etal \cite{ho2020denoising} use a weighted variational objective in practice.

%%%%%%%%%%%%%%%%%%%%%%%%%%%%%
\subsection{Distortion-perception traversal}

The choice to enhance a base codec reconstruction with a generative model has a compelling advantage over approaches that learn to trade off rate, distortion and perception in an end-to-end manner: in theory, it gives us access to an initial reconstruction $\initialrecon$ with maximum fidelity, and an enhanced reconstruction $\recon$ with maximum perceptual quality. 
First, for a perfect encoder and decoder, $\initialrecon$ has the lowest distortion in expectation.
Second, if the encoder and decoder are deterministic, and the enhancement model learns $p(\target|\initialrecon)$ exactly, then we have $p(\recon) = p(\target)$.
This means perfect quality under the definition of Blau and Michali \cite{blau2019rethinking}.
One can also view the decoder and enhancement model as one joint stochastic decoder, which is required for perfect quality at any bitrate \cite{tschannen2018deep}. 
In this picture, the diffusion steps will then gradually move the prediction from the mean of the learned distribution---which would be 0 for the residuals of an optimal base model---to a sample, corresponding to a transition from high fidelity to high perceptual quality.

%%%%%%%%%%%%%%%%%%%%%%%%%%%%%%%%%%%%%%%%%%%%%%%%%%%%%%%
\subsection{Sampling improvements}
\label{sec:method:sampling}

In this work we make use of the noise schedule and sampling procedure introduced by Denoising Diffusion Implicit Models (DDIM) \cite{song2020denoising}.
While we use $T=1000$ diffusion steps during training, the number of sampling steps can be reduced to 100 at test time at negligible cost to performance, by redistributing the timesteps based on the scheme described by Nichol and Dhariwal \cite{nichol2021improved}.
As explained in the previous section, the sampling procedure can be stopped at any point, e.g. when the desired perceptual quality is achieved or when a compute budget is reached. 
To indicate how many sampling steps are performed, we refer to our model as DIRAC-n, going from DIRAC-1 to DIRAC-100.

We further improve sampling efficiency and performance through two contributions: late-start sampling and \emph{rate-dependent thresholding}.

First, we demonstrate in \cref{sec:results:sampling} that we can skip $80\%$ of the 100 sampling steps, instead starting sampling at DIRAC-80 with noise as model input which is scaled according to the diffusion model's noise schedule.
We are not the first to introduce late-start sampling \cite{lyu2022accelerating}, but we can do it while sampling directly from a scaled standard Gaussian as opposed to a more complex distribution, simplifying the approach.
We further explain the effectiveness of this approach by showing that the sampling trajectory has very small curvature in the early steps.

Second, we make use of a method we dub \emph{rate-dependent thresholding}. Like most diffusion works, we scale our data (which here are residuals) to the range $[-1;1]$ and clip all intermediate predictions $\rzero'$ to this range. 
However, the distribution of residuals strongly depends on the bit rate of the base codec, with high rate resulting in small residuals between original and reconstruction, and vice versa (see appendix for an analysis of residual distributions). 
Inspired by Saharia \etal \cite{saharia2022imagen}, who introduce \emph{dynamic thresholding}, we analyze the training data distribution and define a value range for each rate (more precisely, for each $\lambda_{rate}$ we evaluate).
Empirically we found that choosing a range such that $95\%$ of values fall within it works best.
During sampling, intermediate predictions are then clipped to the range for the given rate parameter instead of $[-1;1]$. 
We hypothesize that this reduces outlier values that disproportionately affect PSNR. In \cref{sec:results:sampling} we show that it does indeed improve PSNR, without affecting perceptual quality.
Contrary to Saharia \etal we only perform clipping, but not rescaling of intermediate residuals.

We only apply rate-dependent thresholding in the generative compression setting, where we have access to $\lambda_{rate}$ on the receiver-side, but not for the enhancement of traditional codecs as access to the quality factor is not guaranteed.

\section{Experiments}
\label{sec:experiments}

We evaluate DIRAC both as a generative compression model by enhancing a
strong neural base codec, and in the perceptual enhancement setting by  using traditional codecs as base codec.
In the generative compression setting, we evaluate both rate-distortion and rate-perception performance in comparison to prior work in neural compression, and demonstrate that DIRAC can smoothly traverse the entire rate-distortion-perception tradeoff.
In the enhancement setting, we demonstrate DIRAC's flexibility by comparing it with task-specific methods from the literature,  focusing on both distortion and perceptual quality.
Finally, we present experiments that elucidate why our proposed sampling improvements---extremely late sampling start and rate-dependent thresholding---can be successful in the residual enhancement setting.

%%%%%%%%%%%%%%%%%%%%%%%%%%
\paragraph{Baselines}
\label{sec:experiments:baselines}

For the generative compression setting, we focus on strong GAN-based baselines. 
One of the strongest perceptual codecs is \emph{HiFiC}~\cite{mentzer2020hific}, a GAN-based codec trained for a specific rate-distortion-perception tradeoff point.
\emph{MultiRealism}, a followup work \cite{agustsson2022multi}, allows navigating the distortion-perception tradeoff by sharing decoder weights and conditioning the decoder on the tradeoff parameter~\cite{agustsson2022multi}
Additionally, \emph{MS-ILLM}~\cite{muckley2023improving} show that better discriminator design can further improve perceptual scores \cite{muckley2023improving}.  
Finally, Yang and Mandt~\cite{yang2022lossy} propose a codec where a conditional DDPM takes the role of the decoder, directly producing a reconstruction from a latent variable.

In the JPEG restoration setting, we compare to \emph{DDRM}~\cite{kawar2022jpeg}, which recently outperformed the former state-of-the-art method 
QGAC \cite{ehrlich2020quantization}.
They use a pre-trained image-to-image diffusion model and relax the diffusion process to nonlinear degradation, as introduced in~\cite{kawar2022denoising}, to enable JPEG restoration for low resolution images.

Other relevant diffusion-based baselines include \emph{Palette}~\cite{saharia2022palette} and $\Pi$GDM~\cite{song2023solving}, which explicitly train for JPEG restoration, yet only report perceptual quality on low resolution datasets. 
Finally, for VTM restoration, we consider \emph{ArabicPerceptual}~\cite{wang2022perceptual}, however it is trained and evaluated on different datasets and metrics. 
Due to these differences, comparison to these methods can be found in the appendix.

%%%%%%%%%%%%%%%%%%%%%%%%%%
\paragraph{Our models}
\label{sec:experiments:our_models}

Creating a DIRAC model consists of two stages: (1) training or defining a multi-rate base codec, and (2) training a diffusion model to enhance this base codec.

In the generative compression setting, \ie when the base codec is neural,
we use the SwinT-ChARM \cite{Zhu_Yang_Cohen_2022} model. It is a near-state-of-the-art compressive autoencoder based on the Swin Transformer architecture \cite{liu2021swin}.
We adapt this codec to support multiple bitrates using a technique known as \emph{latent scaling}, see details in appendix.

In the enhancement setting, we couple DIRAC with two standard codecs as base model:
 the intra codec of VTM 17.0 \cite{bross2021overview}, as it is one of the best performing standard codecs in the low bitrate regime,
and the widely-used JPEG \cite{wallace1991jpeg} codec. 
Later, we refer to SwinT-ChARM+DIRAC as just DIRAC, while we explicitly refer to VTM+DIRAC and JPEG+DIRAC.

%%%%%%%%%%%%%%%%%%%%%%%%%%
\paragraph{Metrics and evaluation}
\label{sec:experiments:metrics}

We evaluate our method using both distortion metrics and perceptual quality metrics. 
We always evaluate on full resolution RGB images: we replicate-pad the network input so that all sides are multiple of the total downsampling factor of the network, and crop the output back to the original resolution. We repeat scores as reported in the respective publications.

To measure distortion, we use the common PSNR metric. 
We also include the full-reference LPIPS \cite{zhang2018lpips} metric as it has been shown to align well with human judgment of visual quality. 
As perceptual quality metrics, we primarily use a variation of the Frechet Inception Distance (FID)~\cite{heusel2017fid} metric, which measures the distance between the target distribution $p(\target)$ and the distribution of reconstructions $p(\recon)$. 
FID requires resizing of input images to a fixed resolution, which for high-resolution images will destroy generated details.
We therefore follow procedure of previous compression work and use half-overlapping $256\times256$ crops \cite{mentzer2020hific, agustsson2022multi, muckley2023improving} for high resolution datasets, this metric is referred to as FID/256.

When we report bitrates, we perform entropy coding and take the file size. 
This leads to no more than $0.5\%$ overhead compared to the theoretical bitrate given by the prior. 

\begin{figure*}[t]
    \includegraphics[width=0.93\textwidth]{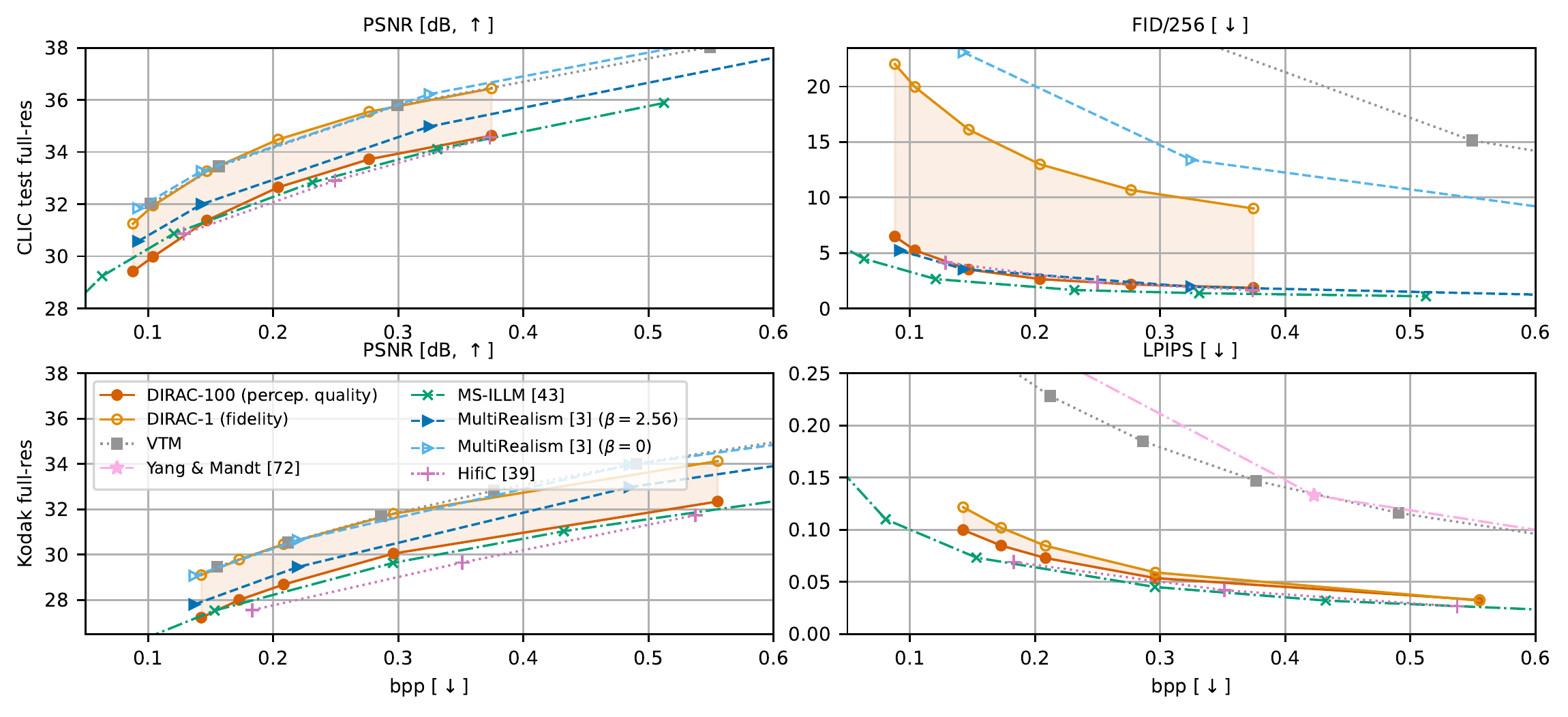}
    \vspace{-2mm}
    \caption{Rate-distortion (left) and rate-perception (right) curves for the CLIC2020 test set (top) and Kodak dataset (bottom).
    The Kodak dataset has too few samples for FID/256 evaluation, instead we evaluate LPIPS, a perceptual distortion metric.
    }
    \label{fig:results:rdp_curves}
\end{figure*}

%%%%%%%%%%%%%%%%%%%%%%%%%%
\paragraph{Datasets} 
\label{sec:experiments:datasets}

To train the SwinT-ChARM base model and residual diffusion models, we use the training split of the high-resolution CLIC2020 dataset~\cite{toderici2020clic} (1633 images of varying resolutions). 
For DDPM training, we follow the three-step preprocessing pipeline of HiFiC~\cite{mentzer2020hific}: for each image, randomly resize according to a scale factor uniformly sampled from the range $[0.5, 1.0]$, then take a random $256 \times 256$ crop, then perform a horizontal flip with probability 0.5.
For validation and model selection, we use the CLIC 2020 validation set (102 images).

We evaluate on two common image compression benchmark datasets: the CLIC2020 test set (428 images) and the Kodak dataset~\cite{kodak} (24 images). To enable comparison with enhancement literature, we evaluate on the low resolution ImageNet-val1k~\cite{deng2009imagenet,pan2021exploiting}. We follow the preprocessing procedure from Kawar \etal \cite{kawar2022jpeg} where images are center cropped along the long edge and then resized to 256.
%%%%%%%%%%%%%%%%%%
\paragraph{Implementation details}
\label{sec:experiments:implementation_details}

For the base codec, we implement SwinT-ChARM as described in the original paper.
We first train a single rate model for 2M iterations on $256\times256$ CLIC 2020 train crops, then finetune it for multiple bitrates for 0.5M iterations.
The standard base codecs are evaluated using the VTM reference software, CompressAI framework \cite{begaint2020compressai} and libjpeg in Pillow.
We provide full details on the implementation and hyperparameters in the appendix.

The diffusion residual model is based off DDPM's official open source implementation \cite{dhariwal2021diffusion},
and we base most of our default architecture settings on the $256 \times 256$ DDPM from Preechakul \etal \cite{preechakul2022diffusion}, which uses a U-Net architecture \cite{unet_ronneberger}. Conditioning on the base codec reconstruction $\initialrecon$ is achieved by concatenating $\initialrecon$ and the DDPM latent $\mathbf{r_t}$ in each step.
Our model has 108.4 million parameters.
For context, the HiFiC baseline has 181.5 million.
As HiFiC requires only one forward pass to create a reconstruction, it is typically less expensive than DIRAC. We provide more detail on computational cost in 
the appendix.

Finally DIRAC and VTM+DIRAC diffusion models are trained for 650k steps, using the Adam optimizer \cite{kingma2014adam} with a learning rate of $10^{-4}$ and no learning rate decay.
The JPEG+DIRAC model was trained for 1M iterations, as JPEG degradations are much more severe than those of VTM and SwinT-ChARM.

\section{Results}
\label{sec:results}

\begin{figure*}[t]
    \centering
    \includegraphics[width=0.98\textwidth]{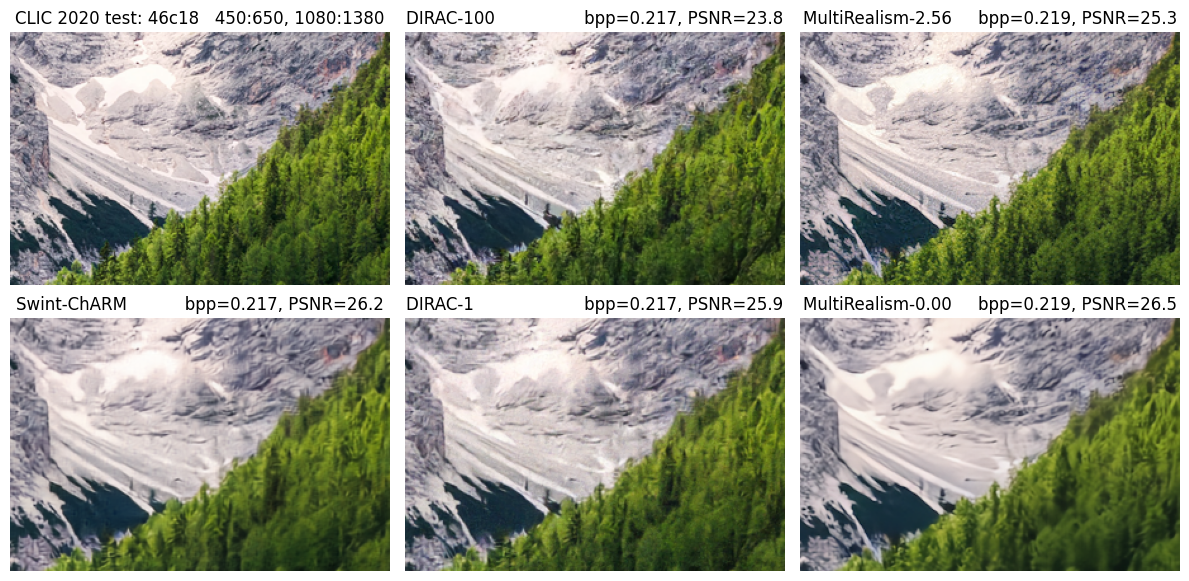}
    \caption{CLIC 2020 test reconstructions comparing our model to \emph{MultiRealism}\cite{agustsson2022multi}. We show original (top left), Swint-ChARM base codec (bottom left), DIRAC-1 (high fidelity) and DIRAC-100 (high perceptual quality) in center column, \emph{MultiRealism} counterparts in right column. Shown scores are for full image. Best viewed electronically.}
    \label{fig:results:qualitative}
\end{figure*}

%%%%%%%%%%%%%%%%%%%%%%%%%%%%%%
\subsection{Generative compression}

We visualize the rate-distortion and rate-perception tradeoffs in \cref{fig:results:rdp_curves}.
We show our model in two configurations: DIRAC-100 (100 sampling steps) has maximum perceptual quality, and DIRAC-1 (single sampling step) has minimal distortion.

Along the distortion axis, \ie PSNR, DIRAC-1 is close to VTM and \emph{MultiRealism}\cite{agustsson2022multi} at $\beta=0$, which in turn is competitive with the state of the art.
On the perceptual quality side, HiFiC is the current state of the art of peer-reviewed works. DIRAC-100 matches HiFiC in FID/256 with better PSNR on both test datasets.
Likewise, we match \emph{MultiRealism} at $\beta=2.56$ in FID/256.
Note that between DIRAC and \emph{Multirealism}, no model is strictly better than the other, \ie better on both distortion and perception axes at the same time.
This is reflected in the examples in \cref{fig:results:qualitative}, where DIRAC-1, our high-fidelity model, looks a bit sharper than \cite{agustsson2022multi}, while examples with high perceptual quality are hardly distinguishable.

\begin{figure*}[t]
    \centering
    \includegraphics[width=0.98\textwidth]{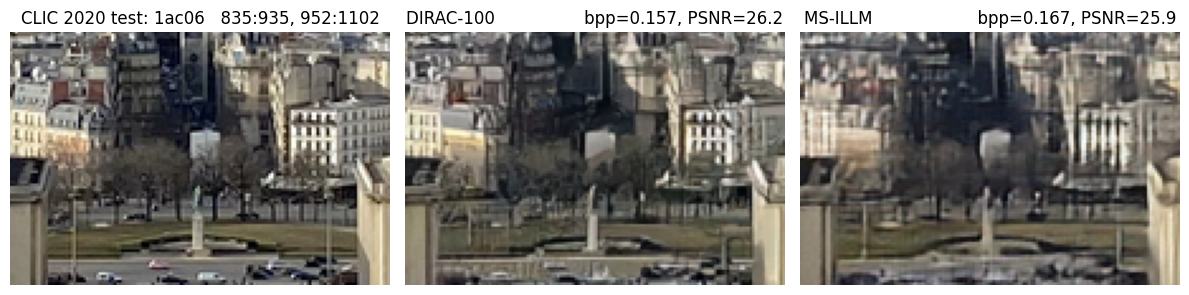}
    \caption{CLIC 2020 test reconstruction by DIRAC-100 and MS-ILLM, crop location chosen based on \cite{muckley2023improving}.}
    \label{fig:results:illm}
\end{figure*}

MS-ILLM \cite{muckley2023improving} achieves a new state of the art in FID/256 and is unmatched by all other methods, but upon qualitative comparison in \cref{fig:results:illm} we observe that even at a lower bitrate compared to MS-ILLM, DIRAC-100 is able to generate perceptually relevant details in a more meaningful manner.
Of course, this is only a single datapoint, and stronger claims require a thorough perceptual comparison.
Similar to Multirealism \cite{agustsson2022multi}, our model can target a wide range of distortion-perception tradeoffs at test time, indicated by the shaded area in \cref{fig:results:rdp_curves}. 
Finally, we compare to the diffusion-based codec of Yang and Mandt \cite{yang2022lossy} on Kodak. 
Our model outperforms theirs by a large margin in terms of LPIPS.
More visual examples can be found in the appendix.

%%%%%%%%%%%%%%%%%%%%%%%%%%%%%%%%%
\subsection{Enhancement of standard codecs}

\begin{figure*}[t]
    \includegraphics[width=\textwidth]{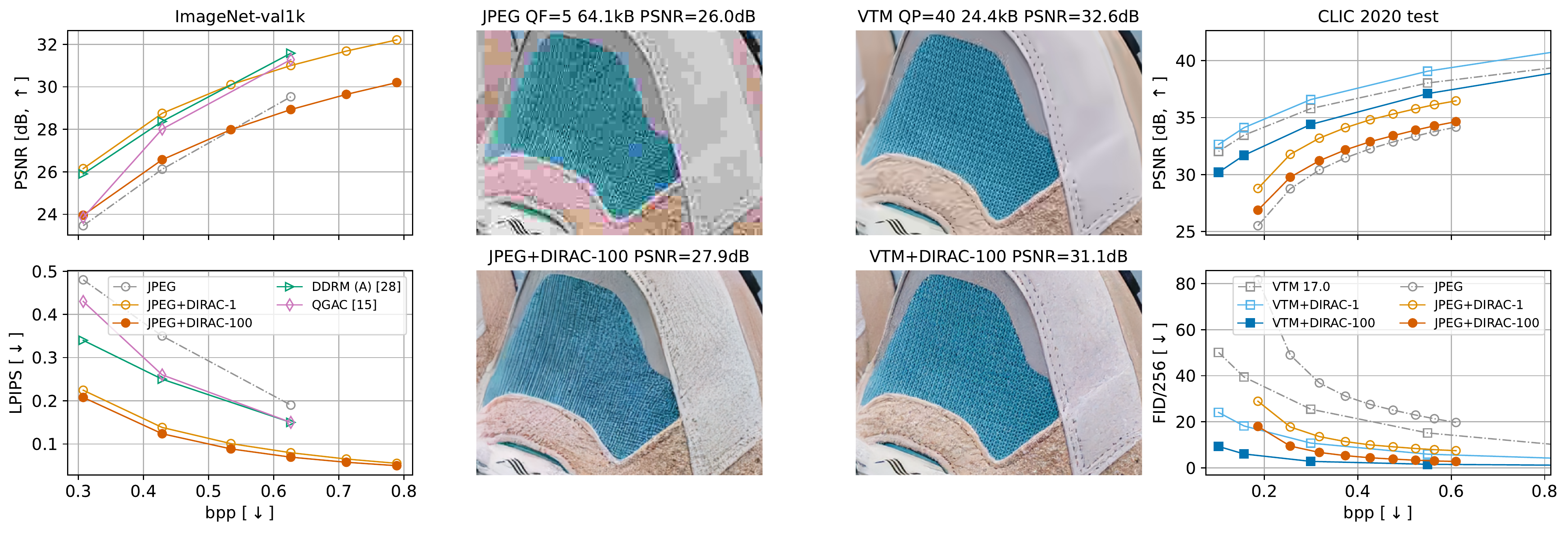}
    \caption{
    Quantitative results for JPEG+ and VTM+DIRAC on ImageNet-val1k (left) and CLIC test 2020 (right) respectively. We show rate-distortion (top) and rate-perception (bottom) curves.
    Qualitative sample is image ``3f273e'' in CLIC 2020 test.
    }
    \label{fig:results:enhancement}
\end{figure*}

\begin{figure*}[t]
    \includegraphics[width=\textwidth]{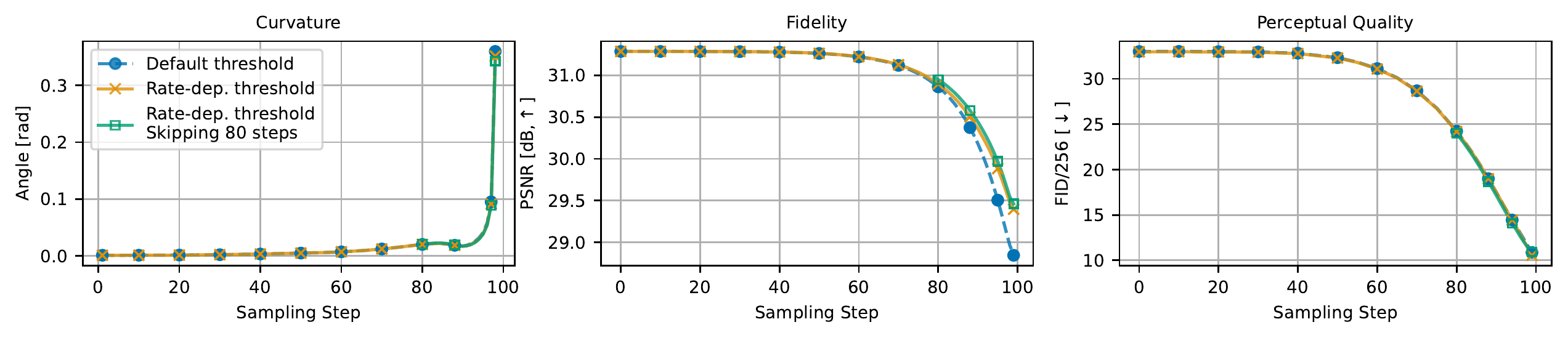}
    \caption{Analysis of the curvature of the sampling trajectory (approximated by the angle between update vectors), as well as the change in PSNR and FID/256 during sampling. All evaluations done on the CLIC 2020 val subset.}
    \label{fig:results:residualsovertime}
\end{figure*}

We evaluate enhancement of two standard codecs: JPEG and VTM. In \cref{fig:results:enhancement} we compare JPEG+DIRAC to literature on the low-resolution dataset ImageNet-1K (left panels) and evaluate JPEG+DIRAC and VTM+DIRAC on the high-resolution dataset CLIC test 2020 (right panels).

When comparing to enhancement literature (left panels in \cref{fig:results:enhancement}), we compare to QGAC and DDRM, specifically their scores resulting from averaging 8 independent samples, denoted DDRM (A). JPEG+DIRAC-1 slightly outperforms the competing methods in the low rate regime in terms of PSNR, while improving LPIPS by a large margin.
Further sampling allows JPEG+DIRAC-100 to improve LPIPS, at the cost of PSNR. While the difference in LPIPS seem small, qualitatively the textures in JPEG+DIRAC-100 are much better than in JPEG+DIRAC-1, as can be seen in the appendix.

When evaluating JPEG+DIRAC and VTM+DIRAC on the high-resolution dataset CLIC test 2020  (right panels in \cref{fig:results:enhancement}), we can see that both VTM+DIRAC-1 and JPEG+DIRAC-1 outperform their base codec in fidelity. In the perceptual enhancement setting, both VTM+DIRAC-100 and JPEG+DIRAC-100 far outperform their base codec in FID/256, specifically at the lowest rate, with a 81\% and 78\% improvement respectively. 
DIRAC offers a consistent boost in perceptual quality, even as one improves the base codec from JPEG to VTM.
We show visual examples of both systems in the middle panels, showing a drastic improvement in visual quality.
Notice the lack of texture on the top samples, which are from distortion-optimized codecs. For VTM, the bottom sample has higher distortion (i.e. lower PSNR), yet looks far better to the human observer.

%%%%%%%%%%%%%%%%%%%%%%%%%%%%
\subsection{Sampling analysis}
\label{sec:results:sampling}

Reverse sampling in diffusion models is equivalent to integrating a stochastic differential equation \cite{Song2020scorebased}.
The error incurred in the numerical approximation of the true solution trajectory will generally be proportional to its curvature \cite{Karras2022elucidating}, meaning parts with low curvature can be integrated with few and large update steps.

In \cref{fig:results:residualsovertime} (left panel) we show the average curvature of sampling trajectories for our model on the CLIC 2020 val dataset, using 100 DDIM steps \cite{song2020denoising}.
Because computing the Hessian is not feasible for the number of dimensions our model operates in, we approximate it with the angle between consecutive update vectors $c=\cos^{-1}(\mathbf{u_t}\mathbf{u_{t-1}}/||\mathbf{u_t}||\cdot||\mathbf{u_{t-1}}||)$, where $u_t\propto(\rzero'(t)-\mathbf{r_t})$ points from the current diffusion latent to the current prediction of the residual.
We find that the curvature is small along a vast majority of steps in the sampling trajectory, meaning it is indeed possible to take a single large update step and only incur a small error.

Moreover, we find that instead of starting from standard normal noise at time $T$ and taking a large integration step to time $t<<T$, it is sufficient to start directly at $t$, using noise as input to the model that is scaled according to the diffusion model's noise schedule.
In our experiments, starting sampling at $t=20$ was a good tradeoff, with final performance almost identical to the full 100 steps (as seen in the center and right panels of \cref{fig:results:residualsovertime}), but saving $80\%$ of required compute.
One might suspect that the above is due to a suboptimal noise schedule (we use the popular \emph{linear} schedule), but we explored several different schedules as well as noise schedule learning \cite{kingma2021variational} and found no improvement in performance.

Besides showing that our model can work efficiently using at most 20 sampling steps, in the generative compression setting we also introduce a concept we call \emph{rate-dependent thresholding}, which we detail in \cref{sec:method:sampling}.
By clipping each intermediate residual prediction to a percentile-range obtained from the training data (we define the range to include $95\%$ of the data at a given rate), we find that we can improve PSNR while not affecting FID.
This can be seen in the center and right panels of \cref{fig:results:residualsovertime}, which also shows how our model performs a smooth traversal between high fidelity (high PSNR) and high perceptual quality (low FID/256).

\section{Discussion and Limitations}

In this work, we propose a new neural image compression method called Diffusion-based Residual Augmentation Codec (DIRAC).
Our approach uses a variable bitrate base codec to transmit an initial reconstruction with high fidelity to the original input, and then uses a diffusion probabilistic model to improve its perceptual quality.
We show that this design choice enables fine control over the rate-distortion-perception tradeoff at test time, which for example enables users to choose if an image should be decoded with high fidelity or high perceptual quality. Paired with a strong neural codec as base model, we can smoothly interpolate between performance that is competitive with the state of the art in either fidelity or perceptual quality. Our model can also work as a receiver-side enhancement model for traditional codecs, drastically improving perceptual quality at sometimes no cost in PSNR. Finally, we demonstrate that our model can work with 20 sampling steps or less, and propose \emph{rate-dependent thresholding}, which improves PSNR of the diffusion model without affecting perceptual quality in the multi-rate setting.

\paragraph{Limitations}
Although our model gives the user control over the amount of hallucinated content, we currently do not control \emph{where} such hallucinations occur.
Similar to GAN-based codecs, we observe that increasing perception sometimes harms fidelity in small regions with semantically important content, such as faces and text.
Addressing this limitation is an important next step for generative codecs.
Additionally, it is fairly expensive to use a DDPM on the receiver side. 
Although we drastically reduce the number of sampling steps, HiFiC and its variations 
\cite{mentzer2020hific,agustsson2022multi} 
are less expensive to run. 
We provide more details on computational cost in the appendix.
On the other hand, sampling efficiency of diffusion models is a major research direction, and we expect our approach to benefit from these advances.

{
\ificcvfinal
    \subsubsection*{Acknowledgments}
    We thank Johann Brehmer, Taco Cohen, Yunfan Zhang, Hoang Le for useful discussions and reviews of early drafts of the paper. Thanks to Fabian Mentzer for instructions on reproducing HiFiC, and to Matthew Muckley for providing the MS-ILLM reconstructions and scores.
\fi
}

%%%%%%%%%%%%%

{
  \small
  \bibliographystyle{ieee_fullname}
  \bibliography{bibliography}
}

%%%%%%%%%%%%%
\newpage

\appendix
\renewcommand\thefigure{\thesection.\arabic{figure}}
\renewcommand\thetable{\thesection.\arabic{table}}

%%%%%%%%%%%%%%%%%%%%%%%%%%%%%%%%%%%%%%%%%%%%%%%%%%%%%%%%%%%%%%%%

%%%%%%%%%%%%%%%%%%%%%%%%%%%%%%%%%%%%%%%%%%%%%%%%%%%%%%%%%%%%%%%%
\section{Method}
\setcounter{figure}{0}
\setcounter{table}{0}

%%%%%%%%%%%%%%%%%%%%%%%%%%%%%%%%%%%%%%%%%%%%%%%%%%%%%%%%%%%%%%%
\subsection{Derivation of DDPM loss} 
\label{appendix:ddpm_derivation}

For completeness, we show the derivation of the objective when a diffusion model learns to predict $\xo$ directly (equivalently, $\rzero$).
This is not a new contribution, and a similar derivation can be found for example in the work of Nichol and Dhariwal \cite{nichol2021improved}.

Following Sohl-Dickstein \etal \cite{sohl2015deep}, the ELBO on the log likelihood of $p_\theta(\xo)$ can be written as $T$-step Kullback–Leibler (KL) divergences: $KL(q(\xpast|\xt, \xo) || p_\theta(\xpast|\xt))$ for $t = 1, ..., T-1$. 
Under the assumption that the two distributions of interest are both Gaussian, we further assume $\Sigma_\theta(\xt, t)=\sigma^2_t \mathbf{I}$ to be time dependent constants. 
This reduces the KL divergence at step $t$ to a comparison from the model mean to the posterior mean of the forward process:
\begin{align} \label{eq:kl}
    KL(q(\xpast|\xt, \xo) || p_\theta(\xpast|\xt)) = \quad \quad \quad \\
       \quad \quad \E_q \bigg[ \frac{1}{2\sigma^2_t} || \tmut (\xt, \xo) - \mutheta(\xt, t) ||^2 \bigg] + C, 
\end{align}
where $C$ is a constant, and $\tmut$ denotes the posterior mean of the forward process conditioned on the data input $\xo$, i.e., $q(\xpast|\xt, \xo)$. 

Note that, using the Bayes' rule and the Markovian property of the forward process, we can rewrite
\begin{align}
   \nonumber q(\xpast|\xt, \xo) & = \frac{q(\xpast, \xt|\xo)}{q(\xt | \xo)} \\
   & = \frac{q(\xpast|\xo) q(\xt|\xpast)}{q(\xt | \xo)}, \label{eq:qxtx0}
\end{align}
where the distribution $q(\xt | \xo) = \N (\xt; \sqrt{\bat} \xo, (1 - \bat) \mathbf{I})$ with $\alpha_t := 1 - \beta_t $ and $\bat := \prod_{s=1}^t \alpha_s$. The $\alpha_t/\beta_t$ are typically chosen empirically and referred to as the \emph{noise schedule}.

With the known distribution of $q(\xt | \xo)$, we can sample $\xt$ at an arbitrary step $t$:
\begin{align} \label{eq:forward_xt}
    \xt(\xo, \e) = \sqrt{\bat} \xo + \sqrt{1-\bat} \e, \quad\quad \e \sim \N(\zero, \mathbf{I}). 
\end{align}

Now each of the components on the RHS of \cref{eq:qxtx0} is defined as a Gaussian distribution with known parameters.  As a result, we can find the posterior mean $\tmut$ in the following explicit form:
\begin{align} \label{eq:mut}
    \tmut(\xt, \xo) 
        &:= \frac{\sqrt{\batminus} \beta_t}{1 - \bat} \xo + \frac{\sqrt{\alpha_t} (1 - \batminus)}{1 - \bat} \xt \\ 
        &= \eta_t\xo+\xi_t\xt\;.
\end{align}

By plugging \cref{eq:mut} into \cref{eq:kl}, and choosing a parametrization that matches $\mutheta$ to $\tmut$ at each diffusion step $t$, for example $\mutheta(\xt, t) = \eta_t g_\theta(\xt, t) + \xi_t \xt$ (where $g_\theta$ is a function that directly predicts $\xo$ from $\xt$), we arrive at an objective function at diffusion step $t$:
\begin{align}
    \min \mathop{\E}_{t,\xo, \e} \bigg[w_t ||\xo - g_\theta(\xt, t) ||^2 \bigg] \quad \text{for} \quad t = 1, ..., T-1. 
     \label{eq:appendix:ddpmloss}
\end{align}
where $w_t := \eta_t^2 / 2 \sigma^2_t$ and $t$ is absorbed into the expectation instead of being summed over.

%%%%%%%%%%%%%%%%%%%%%%%%%%%%%%%%%%%%%%%%%%%%%%%%%%%%%%%%%%%%%%%
\subsection{Noise schedule and loss weights}
\label{appendix:noise_schedule}

\begin{figure}[t]
    \centering
    \includegraphics[width=\linewidth]{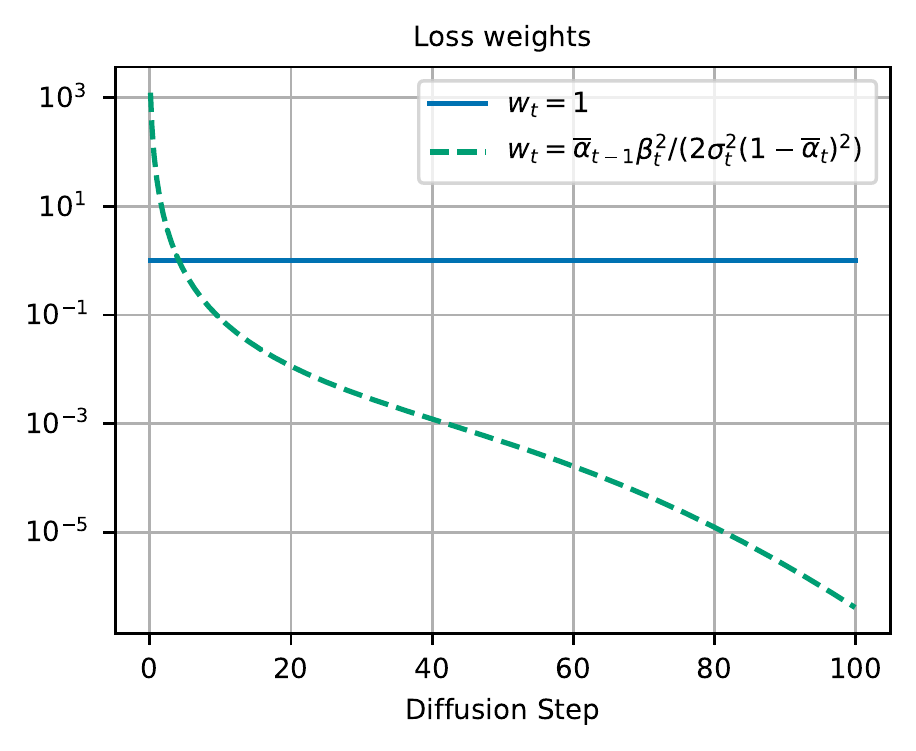}
    \caption{Comparison of magnitudes for the theoretically derived loss weights and our reweighting ($w_t=1$).}
    \label{fig:appendix:loss_weights}
\end{figure}

We work with the so-called \emph{linear} noise schedule \cite{ho2020denoising}, defined as:
\begin{equation}
    \beta_t = (T-t)/(T-1) \cdot \beta_1+(t-1)/(T-1) \cdot \beta_T,
\end{equation}
where $t=1,\ldots,T$, $\beta_1=10^{-4}$ and $\beta_T=0.02$. We tried several different ones, and found that this offered the best combination of PSNR and FID/256. We provide an ablation in \cref{appendix:noise_schedule_ablation}.

We reweight the individual loss terms from \cref{eq:appendix:ddpmloss} to $ w_t=1 $, inspired by the choice of Ho \etal \cite{ho2020denoising}, who do the same for the loss formulation that predicts the noise instead of the denoised sample.
The theoretically derived $w_t=\eta^2/(2\sigma_t^2)$ result in extremely large weights for small $t$, as seen in \cref{fig:appendix:loss_weights}.
We find that $w_t=1$ works better, likely because of the more evenly distributed loss scales, but it is possible that better weightings exist.
Note that the reweighting of the $\epsilon$-objective of Ho \etal results in different weighting than $w_t=1$ due to the reparametrization.

%%%%%%%%%%%%%%%%%%%%%%%%%%%%%%%%%%%%%%%%%%%%%%%%%%%%%%%%%%%%%%%
\subsection{Motivation for modeling residuals}
\label{appendix:ddpm_residual}

\begin{figure*}[t!]
    \centering
    \includegraphics[width=0.8\textwidth]{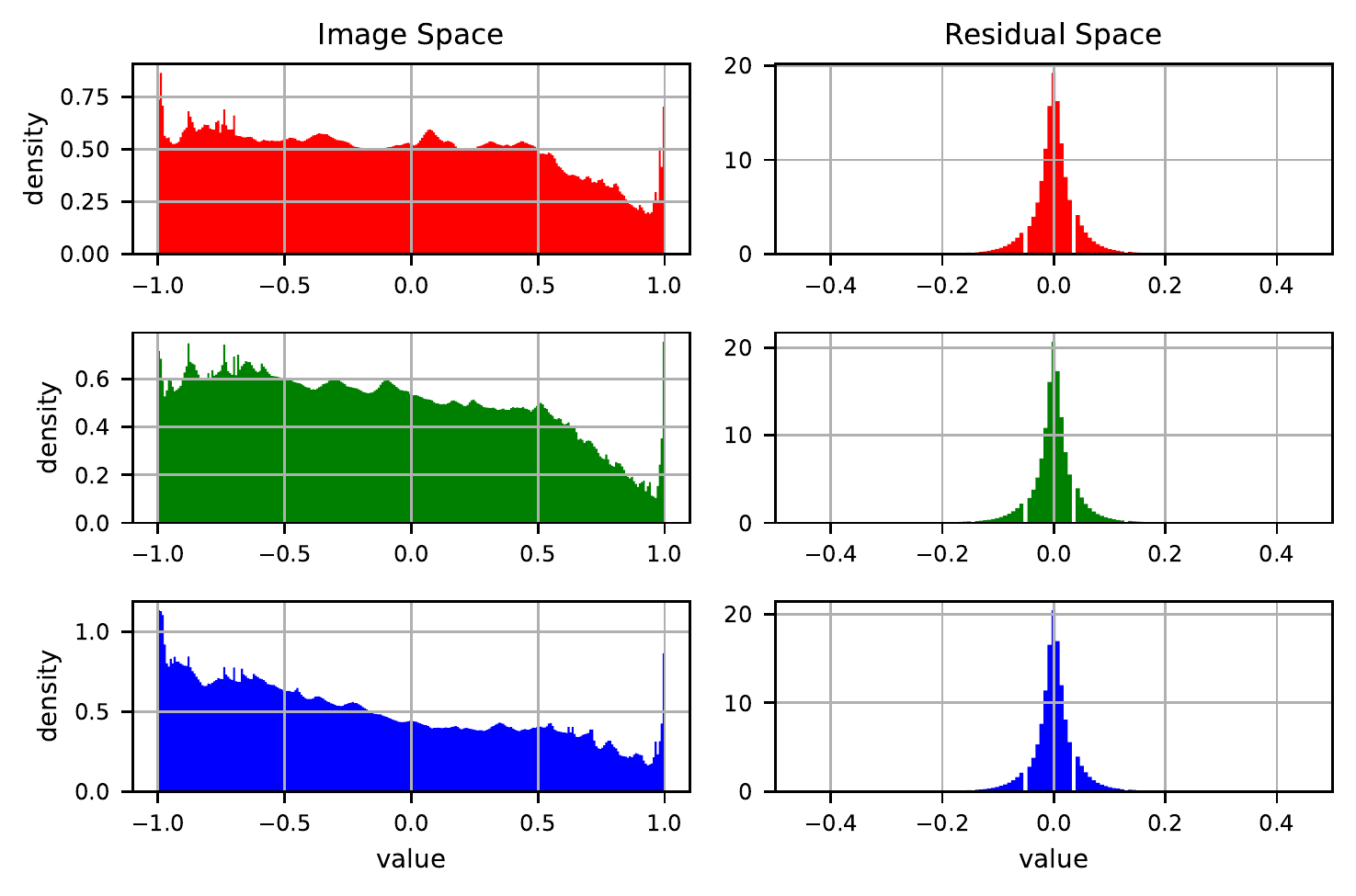}
    \caption{
        Distributions of pixel values in a 1000 random 256x256 crops of the CLIC train dataset, left for the target image $\target$, right for the residual $\residual = \target - \initialrecon$ where the initial reconstruction $\initialrecon$ is from our SwinT-ChARM base model, using random $\lambda_{\text{rate}}$ values.
        Rows corresponds to the red, green and blue channels respectively.
    }
    \label{fig:appendix:methods:pixel-distribution-clic-train}
\end{figure*}

DIRAC outputs an initial reconstruction $\initialrecon$, then models the distribution of residuals $p(\residual|\initialrecon)$ using a DDPM, where $\residual = \target - \initialrecon$ (we drop the index 0, which denotes the data space for our diffusion model). 
In theory, one could model the image distribution $p(\target|\initialrecon)$ directly. 
From an entropy perspective, the two approaches are equal: the entropy of $p(\residual|\initialrecon)$ is equal to the entropy of $p(\target|\initialrecon)$ since, given the data $\target$, the residual $\residual$ is a deterministic function of the initial reconstruction.
Yet, while their entropy might be similar, the shape of these distributions is different.

In \cref{fig:appendix:methods:pixel-distribution-clic-train}, we show the histogram of pixel values in the target images (left) and the residuals (right) for each RGB channel, for randomly varying rate factors $\lambda_\text{rate}$. 
We observe that residual values approximately follow a normal distribution, whereas the pixel values in image space show a more intricate distribution. 
Given the fact that DDPMs (in their typical formulation) map Gaussian noise to the target distribution, we conjecture that it is desirable that the target distribution is close to a normal distribution, and that it helps reduce the number of sampling steps required to obtain satisfactory perceptual quality. Moreover, the residuals never exceed the bounds of image values, i.e. $[-1,1]$, so that the typical clipping to this range during sampling has no effect. As a result, we introduce rate-dependent thresholding, which we explain in more detail in \cref{appendix:ddim_sampling}.

%%%%%%%%%%%%%%%%%%%%%%%%%%%%%%%%%%%%%%%%%%%%%%%%%%%%%%%%%%%%%%%
\subsection{Details on sampling procedure}
\label{appendix:ddim_sampling}

We use this section to provide more details on the sampling in our diffusion. Specifically, we describe 1) the early stopping that results in a smooth transition between high fidelity and high perceptual quality, 2) the late-start sampling, which allows us to skip many more steps than earlier works \cite{lyu2022accelerating}, and 3) the \emph{rate-dependent thresholding} we propose.
The observed speedup in sampling corroborates earlier findings \cite{ho2020denoising, nichol2021improved}.

%%%%%%%%%%%%%%%%%%%%%%%%%%
\paragraph{Early stopping}

The DDPM enhancement model of DIRAC predicts the (often sparse) residual $\residual_0 = \target - \initialrecon$.
We find that we are able to stop sampling at any point before we reach the final step, by simply using the intermediate prediction for the residual $\mathbf{r}'_0(t) = g_\theta(\mathbf{r}_t, t)$ as the final sample. 
Specifically, we observe that during sampling, early intermediate predictions are close to zero, and that they become sharper over time.
Stopping early then typically means higher PSNR, whereas stopping late results in high perceptual quality.
Intuitively, this makes sense as well: we know that in the limit of perfect models, $\initialrecon$ has the highest possible expected fidelity, and the DDPM can only increase quality by decreasing fidelity.

Stopping early in this manner is somewhat similar to the scheme proposed by Ho \etal \cite{ho2020denoising}, who use this intermediate prediction to describe a progressive coding scheme.
However, we have already transmitted the initial reconstruction $\initialrecon$, and all DDPM sampling happens on the receiver side, so the settings are quite different. 
It is the receiver-side generation of residuals that allows to navigate the the distortion-perception tradeoff \cite{blau2019rethinking}, and that enables early stopping to achieve a desired tradeoff or to reduce compute requirements.
Optimizing sampling trajectories of diffusion models is an active field, and further research on the sampling in different (conditional) settings may lead to actionable insights \cite{deja2022analyzing}.

%%%%%%%%%%%%%%%%%%%%%%%%%%
\paragraph{Late start}

Similar to findings of \cite{nichol2021improved} and \cite{lyu2022accelerating}, we observe that it is possible to skip several sampling steps.
In particular, we take an initial noise sample and scale it to match the expected standard deviation at timestep $t$ given by the forward process, then plug this ``latent'' into the reverse process. In our case this means $\mathbf{r}_t \sim \mathcal{N}(0, \sigma_t^21)$.
Nichol \etal \cite{nichol2021improved} mainly use the late-start observation to motivate the use of a different noise schedule. 
Based on the similarity of our observations to theirs, it is possible that there are noise schedules that are better suited for the image compression setting, but we tried several different ones and found that our choice worked best out of the ones we tested (see \cref{appendix:noise_schedule_ablation}). 
The key finding in our work is that we can skip a large part (up to 80\%) of the initial steps without performance degradation.

%%%%%%%%%%%%%%%%%%%%%%%%%%
\paragraph{Rate-dependent thresholding}

Typically, diffusion works clip the latents to the $[-1,1]$ as it corresponds to the normalized image-space. Yet we observe that both ground-truth and DIRAC-predicted residuals occupy a smaller value range than $[-1,1]$. We seek to make better use of clipping, and adjust it to the data range at hand. However, early experiments adjusting the clipping thresholds to a single smaller value---based on range percentiles from the training data---did not improve performance. Instead, we set rate-dependent thresholds for clipping, choosing the thresholds such that $95\%$ of residuals in the training data fall within that range, at the given rate factor $\lambda_\text{rate}$.

We perform this analysis for 20 different $\lambda_\text{rate}$, and the resulting thresholds range from $0.0706$ at the highest rate to $0.1490$ at the lowest rate. At test time, if the desired $\lambda_\text{rate}$ is not in the available set, we use the thresholds from its nearest neighbor.

%%%%%%%%%%%%%%%%%%%%%%%%%%%%%%%%%%%%%%%%%%%%%%%%%%%%%%%%%%%%%%%%%%%%%%%%%%%%%%%
\section{Implementation details}
\label{appendix:implementation}
\setcounter{figure}{0}
\setcounter{table}{0}

\subsection{Reproducibility}
\label{appendix:implementation:reproducibility}

 %%%%%%%%%%%%%%%%
\paragraph{Datasets}

All datasets used in this work are publicly available.
We show our three-step data augmentation pipeline for DDPM training in Section 4.1 under Datasets. 

To compare to Kawar \etal (DDRM) \cite{kawar2022jpeg} and QGAC \cite{ehrlich2020quantization}, we use the Imagenet val-1k dataset.
The Imagenet val-1k set consists of the first image from each class in the validation set, alphabetically ordered, as reported in \cite{pan2021exploiting}.
The exact filenames can also be found in 
\href{https://github.com/XingangPan/deep-generative-prior/blob/master/scripts/imagenet_val_1k.txt}{the corresponding Github repo}.
For direct comparison to Kawar \etal \cite{kawar2022jpeg}, we use the data augmentation procedure shown in \href{https://github.com/bahjat-kawar/ddrm-jpeg/blob/master/datasets/imagenet_subset.py}{their Github repo} during evaluation: a square center crop the size of the shortest dimension, followed by a resizing to $256\times256$.  

To compare to Palette \cite{saharia2022palette} and $\Pi$GDM \cite{song2023solving} in \cref{appendix:additional_enhancement_comparison}, we use the Imagenet ctest10k dataset.
The Imagenet ctest10k subset consists of 10,000 images from the Imagenet validation set, as originally reported by \cite{larsson2016learning}.
The original page is no longer online, but a copy of the filenames can be found on \href{https://github.com/zhaoyuzhi/Reference-Based-Sketch-Image-Colorization-ImageNet/blob/main/util/ctest10k.txt}{Github}.
We use PIL to resize each image so that the shortest side is of size $256$, then perform a $256\times256$ center crop. 

Finally, to compare to ArabicaPerceptual \cite{wang2022perceptual} in \cref{appendix:additional_enhancement_comparison}, we use the CLIC 2022 validation dataset \cite{toderici2020clic}, which is comprised of 30 high-resolution images. Note that due to its small size, it makes it unreliable for FID/256, hence we only report LPIPS as proxy for a perceptual metric.

%%%%%%%%%%%%%%%%
\paragraph{Models}

The base neural image codec is a variation on a mean-scale hyperprior \cite{balle2018variational} called the SwinT-ChARM hyperprior \cite{Zhu_Yang_Cohen_2022}.
High quality implementations of neural image codecs are available via \href{https://github.com/InterDigitalInc/CompressAI}{CompressAI} \cite{begaint2020compressai}, see for example this 
\href{https://github.com/InterDigitalInc/CompressAI/blob/master/compressai/models/google.py}{mean-scale hyperprior implementation link on GitHub}.
This library also provides entropy coding functionality.
A reproduced version of the SwinT-ChARM model is \href{https://github.com/Nikolai10/SwinT-ChARM}{provided on GitHub by user Nikolai10}.
We provide more details on the neural base codec in \cref{appendix:implementation:neuralbasecodec}.

Reconstructions for standard codecs were obtained using CompressAI \cite{begaint2020compressai}.
The commands to reproduce these are given in \cref{appendix:implementation:standardbasecodec}.

The DDPM component was trained using the open source implementation of \cite{dhariwal2021diffusion}. 
The main change in implementation is that our DDPM is conditioned on an initial reconstruction from a mean-scale hyperprior, which is achieved by concatenating it with the DDPM latent (these two tensors have the same spatial dimensions).
We provide information about hyperparameters in \cref{appendix:implementation:dirac}.
Lastly, we provide information about computational complexity and training compute in \cref{appendix:implementation:compute}.

%%%%%%%%%%%%%%%%%%%%%%%%%%
\begin{table}[t]
    \caption{Hyperparmeters at each stage of training.}
    \label{tab:appendix:implementation:training}
    \centering
    \begin{tabular}{l r r r}
      \toprule
                                  & Single rate & Multi-rate & \\
         Parameter                & SwinT & SwinT & DDPM \\
       \midrule
         Steps                    & 2M         & 500k       & 650k   \\
         Learning rate            & 1e-4       & 1e-4       & 1e-4   \\ 
         Batch size               & 8          & 8          & 64     \\
         $\lambda_\textbf{rate}$  & 0.0016     & random     & random \\ 
         $\lambda_\textbf{LPIPS}$ & 0.0065     & 0.0065     & 0.001  \\ 
       \bottomrule
    \end{tabular}
\end{table}
%%%%%%%%%%%%%%%%%%%%%%%%%%

%%%%%%%%%%%%%%%%%%%%%%%%%%
\subsection{Neural base codec} 
\label{appendix:implementation:neuralbasecodec}

Training a neural base codec is a two stage process: we first train a base model for a single bitrate for 2M iterations, then finetune it to operate under multiple bitrates for 500k iterations.
Hyperparameters for these two stages are shown in \cref{tab:appendix:implementation:training}.

%%%%%%%%%%%%%%%%%%%%%%%%%%
\begin{table}[t]
    \caption{Hyperparameters for our SwinT-ChARM model.}
    \label{tab:appendix:implementation:basecodec}
    \centering
    \begin{tabular}{l r r}
    \toprule
        Parameter & Enc/Dec & Hyper Enc/Dec \\
    \midrule
        Patch size    & 2         & 2 \\
        Embed dim     & 64        & 64 \\
        Window size   & 8         & 4 \\
        Blocks        & [2,2,6,2] & [4,2] \\
        Head dims     & 32        & 32 \\
        Normalization & LayerNorm & LayerNorm \\
        Code channels & 320       & 192 \\
    \bottomrule
    \end{tabular}
\end{table}
%%%%%%%%%%%%%%%%%%%%%%%%%%

%%%%%%%%%%%%%%%%%%%%%%%%%%
\paragraph{Single rate SwinT-ChARM}

We use the ``SwinT-ChARM hyperprior'' architecture of Zhu \etal \cite{Zhu_Yang_Cohen_2022} as neural base codec.
This is a hierarchical VAE with quantized latent variables, similar to the mean-scale hyperprior of Ball\'e \etal \cite{balle2018variational}.
The encoder and decoder networks are built using Swin Transformers \cite{liu2021swin}.
The first level encoder and decoder produce the quantized latent variable and decode it to a reconstruction.
The second level, which is the prior model, uses a hyper-encoder to produce a so-called quantized hyper-latent $\rvz$, then maps that to parameters $\mu_\quantlatent, \sigma_\quantlatent$ using a hyper-decoder.
The hyper-latent distribution is modeled using an unconditional prior $p(\rvz)$ \cite{balle2018variational}, so that the hyper-latent can be transmitted losslessly using entropy coding.
The probability of the quantized latent under the prior is then equal to $ p( \quantlatent | \rvz ) = \mathcal{N}( \quantlatent | \mu_\quantlatent, \sigma_\quantlatent) $.

The encoder and decoder contain transformer blocks at four resolutions, the hyper-encoder and hyper-decoder at two resolutions.
Special care must be taken to ensure that the latent resolution is a multiple of the hyper codec window size.
To enable transmission of images that result in latents with spatial dimensions not divisible by this factor, we use `replicate' padding to pad the image, transmit the padded image, then crop to the original resolution on the receiver side.
We transmit the original spatial dimensions as 16 bit integers. 
This bit cost is negligible compared to the cost of transmitting the content.

We use 320 channels for every layer in the encoder/decoder, and 192 channels for the hyper-encoder/hyper-decoder.
Latent quantization is performed using a ``mixed'' strategy: both decoders see the quantized latents during training, and a straight-through estimator is used to make sure gradients pass through the hard quantization operation; the prior computes the loss based on latents quantized using additive uniform quantization noise $\rvu \sim U(-0.5, 0.5)$ \cite{guo2021soft}.
For more details and architecture visualizations, we refer the reader to \cite{Zhu_Yang_Cohen_2022, liu2021swin}.
Our used hyperparameters are listed in \cref{tab:appendix:implementation:basecodec}.

%%%%%%%%%%%%%%%%%%%%%%%%%%%%
\paragraph{Multi-rate SwinT-ChARM}
\label{appendix:implementation:multiratebasecodec}

To obtain a multi-rate base codec, we first train a single rate SwinT-ChARM hyperprior for 2 million iterations using $\lambda_\text{rate} = 0.0016$.
Multi-rate capabilities are added to this model using a technique known as latent scaling \cite{chen2020variable, plonq}.
This procedure effectively changes the latent quantization binwidth by \emph{scaling the latent variable} according to the tradeoff parameter $\ratetradeoff$.

We achieve this in practice by mapping $\ratetradeoff$ to a scaling value $s$ using an exponential map:
\begin{equation}
\label{eq:swint_exp_map}
    s = \alpha_s \exp( \beta_s \cdot \log \ratetradeoff )
      = \alpha_s (\ratetradeoff)^{\beta_s},
\end{equation}
\noindent
where $\alpha_s$ and $\beta_s$ are learnable parameters.
Other parametrizations are possible too, this parametrization has the advantage that $s$ is positive if $\alpha_s \geq 0$, thus avoiding instability.
The unquantized latent is multiplied by this quantized scalar $s$ before being passed to the prior and quantization.
The scaling value $s$ is transmitted as a 16 bit integer at negligible cost.
On the receiver side, the latent is divided by $s$ after decompression, before being passed to the decoder.

Enabling the model to operate under different bitrates, and learning $\alpha_s$ and $\beta_s$, then requires that we sample different $\lambda_\text{rate}$ during training.
For each training batch, we sample a value $\lambda' \sim U[0, 1]$.
The final sampled tradeoff parameter $\ratetradeoff$ is then obtained via interpolation in log space between a pre-specified minimum and maximum:
\begin{equation}
   \log_2 \ratetradeoff = (\log_2(\lambda_\text{max}) - \log_2( \lambda_\text{min})) \times \lambda' + \log_2(\lambda_\text{min})\;,
   \label{eq:lambda_sampling}
\end{equation}
\noindent where we choose $\lambda_\text{max}=0.0160$ and $\lambda_\text{min}=0.0004$. 
Given $\ratetradeoff$, the multirate codec---which now includes the original single rate model and the parameters $\alpha_s, \beta_s$---is trained using a rate-distortion loss.

At test time, the user picks the rate-distortion operating point by selecting a $\lambda_\text{rate}$ value.
The mapping in \cref{eq:swint_exp_map} is used to get the corresponding scalar $s$ for latent scaling.
We find in practice that latent scaling, when compared to schemes similar to the one-hot conditioning of \cite{song2021variable, rippel2021elfvc}, performs slightly better near scale $s = 1$, and slightly worse at the extreme bitrates.

%%%%%%%%%%%%%%%%%%%%%%%%%%
\subsection{Standard base codecs} 
\label{appendix:implementation:standardbasecodec}

%%%%%%%%%%%%%%%%%%%%%%%
\paragraph{VTM base model}

VTM is the reference implementation of the VVC standard.
We run \href{https://vcgit.hhi.fraunhofer.de/jvet/VVCSoftware_VTM/-/releases/VTM-17.0}{VTM-17.0}, and use CompressAI \cite{begaint2020compressai}  to prepare the encoding command.
CompressAI converts given input RGB images to YUV444 before coding them in ``all intra'' mode, then converts the reconstructed YUV444 images back to RGB.
These conversions are lossless.
We use the default all intra configuration, and use QPs $\{ 22, 27, 32, 37, 40 \}$, where higher QPs corresponds to low bitrate and vice versa.

Specifically, let \texttt{\$VTM} be the path to the VTM-17.0 folder, and \texttt{\$IMAGEFOLDER} be the path to an input image folder.
The command used to gather VTM-17.0 evaluations is then:
{
\small
\begin{verbatim}
  python -m compressai.utils.bench vtm 
  $IMAGEFOLDER
  -c $VTM/cfg/encoder_intra_vtm.cfg 
  -b $VTM/bin 
  -q [22, 27, 32, 37, 40]
\end{verbatim}
}

%%%%%%%%%%%%%%%%%%%%%%%
\paragraph{JPEG base model}

JPEG is a well-known standard image codec.
To produce JPEG reconstructions and bitstreams, we use the JPEG functionality in Pillow. 
Similar to VTM, we use CompressAI to prepare the encoding command for multiple quality factors:
{
\small
\begin{verbatim}
  python3 -m compressai.utils.bench jpeg 
  $IMAGEFOLDER
  -q [5, 10, 15, 20, 25, 30, 35, ..., 95] 
\end{verbatim}
}

%%%%%%%%%%%%%%%%%%%%%%%
\subsection{DIRAC}
\label{appendix:implementation:dirac}

Assume a multirate base image codec is available.
We train the DDPM to enhance the reconstructions $\initialrecon$ by conditioning on this image-to-enhance.
In order to support multiple bitrates, the DDPM needs to see reconstructions for many different $\lambda_\text{rate}$ at training time.
In practice, this can be achieved by using the $\lambda_\text{rate}$ sampling technique used for multi-rate SwinT-Charm hyperprior training, meaning the DDPM will see reconstructions with random compression rates during training.
For the VTM and JPEG base models, we follow a similar procedure to sample $\ratetradeoff$, and discretize it to the integer grid so that it can be used as QP or quality parameter.
The DDPM is trained for 650,000 iterations, see the hyperparameter settings specified in \cref{tab:appendix:implementation:training}. It is not conditioned on $\lambda_\text{rate}$, as we found it to have no additional benefit. 

% \guillaume{todo: add description of how multirate is handled with VTM and JPEG.}

Many of our U-Net architecture choices were adopted from the open-source implementation of \cite{dhariwal2021diffusion}, and we refer the reader to  \cite{nichol2021improved, dhariwal2021diffusion} for a detailed explanation of all hyperparameters.
Table \ref{tab:appendix:implementation:architecture} provides an overview of the hyperparameter settings for the DIRAC model.
Horizontal lines separate U-Net parameters and diffusion process parameters.

The channel multiplier corresponds to the increase in width with respect to the base number of channels in each layer.
Following \cite{preechakul2022diffusion}, we train our largest models using a U-Net with 6 separate blocks, increasing the multiplier every 2 blocks.
Self-attention layers are removed from all layers except the bottleneck of the U-Net.
We use the linear noise schedule of \cite{nichol2021improved}, as we saw better performance with this schedule in early experiments, see also \cref{appendix:noise_schedule_ablation} for results with different noise schedules.

\begin{table*}[h!]
    \caption{Hyperparameters for DIRAC during training and sampling. We refer the reader to the official open source implementation of Nichol and Dhariwal \cite{nichol2021improved} for more details on these parameters.}
    \label{tab:appendix:implementation:architecture}
    \centering
    \begin{tabular}{l r r}
     \toprule
     Parameter   & Command line        & DIRAC parameter value \\
     \midrule
     % sorry, messed up the table                           
         Channel multiplier  & \texttt{channel\_mult}     & 1,1,2,2,4,4  \\
         Base num. channels  &  \texttt{num\_channels}     & 128           \\
         Learn $\sigma_t$  & \texttt{learn\_sigma}     & false         \\
         Group normalization   & \texttt{group\_norm}       & true          \\
         Attention resolutions   & \texttt{attention\_} & none     \\
         Num. attention heads   & \texttt{num\_heads}        & 1             \\
         Objective             & \texttt{predict\_xstart} & true          \\
         Num. ResBlocks      & \texttt{num\_res\_blocks}  & 2             \\
         Scale shift conditioning & \texttt{use\_scale\_shift\_norm}   &    false \\
     \midrule                                                            
         Diffusion steps       & \texttt{diffusion\_steps}   & 1000         \\
         Noise schedule        & \texttt{noise\_schedule}    & linear       \\
         Use DDIM sampling      & \texttt{use\_ddim}          & true         \\
         Resampling timesteps   & \texttt{timestep\_respacing}& ddim100       \\
     
    \end{tabular}
\end{table*}

%%%%%%%%%%%%%%%%%%%%%%%%%%%%%%%%%%%%%%%%%%%%%%%%%%%%%%%%%%%%%%%
\section{Results}
\label{appendix:additional_results}
\setcounter{figure}{0}
\setcounter{table}{0}

%%%%%%%%%%%%%%%%%%%%%%%%%%%%%%%%%%%%%%%%%%%%%%%%%%%%%%%%%%%%%%%
\subsection{Additional metrics}
\label{appendix:additional_metrics}

\begin{figure*}[t]
    \centering
    \includegraphics[width=0.9\textwidth]{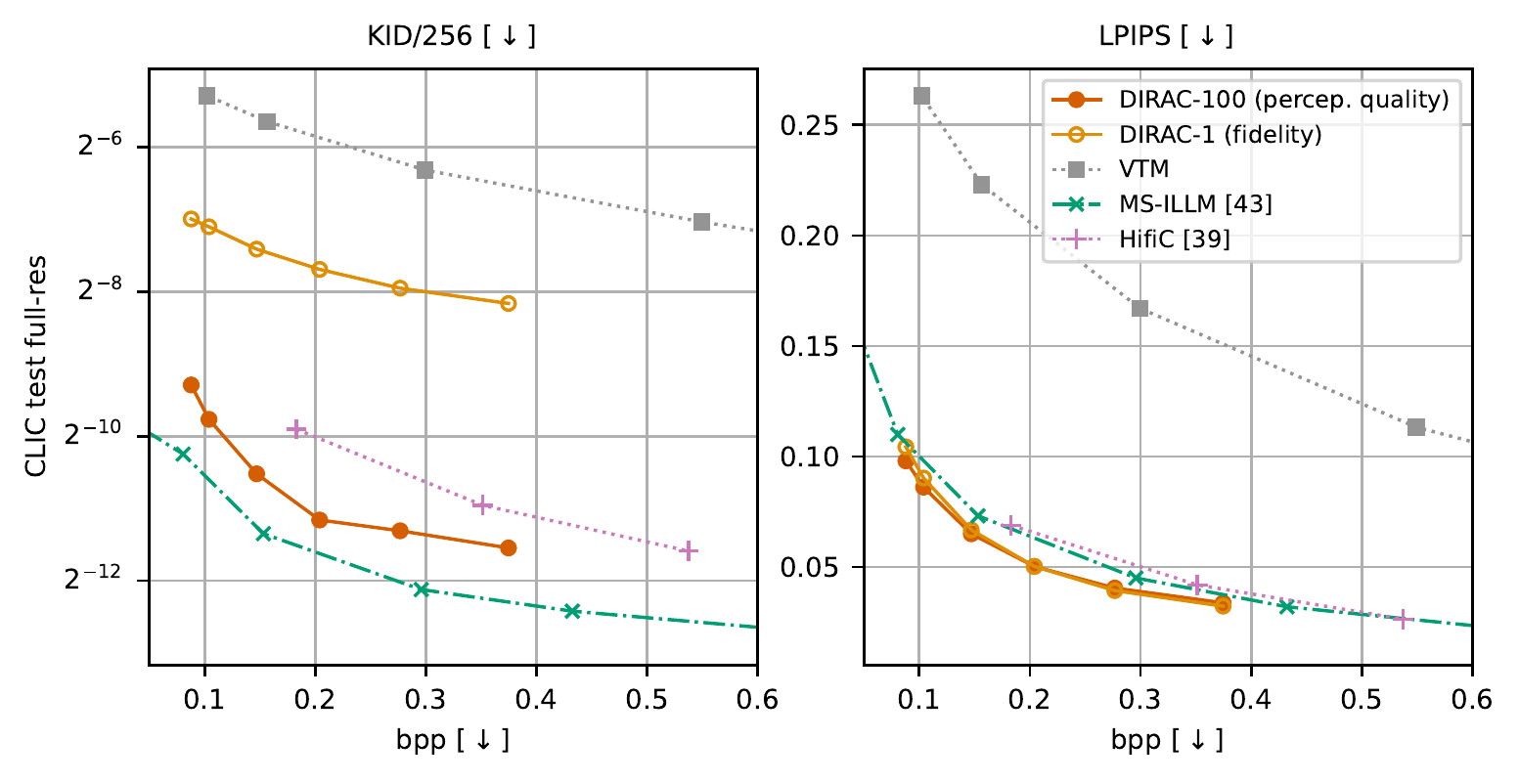}
    \caption{Additional perceptual and distortion metrics for the CLIC 2020 test set.
    }
    \label{fig:appendix:rp_curves}
\end{figure*}

In \cref{fig:appendix:rp_curves} we show additional metrics for DIRAC on the CLIC 2020 test dataset, namely Kernel Inception Distance (KID) \cite{binkowski2018demystifying} and LPIPS \cite{mittal2013niqe}. 
KID is a distribution metric and thus not very reliable on Kodak, and we therefore only show metrics for CLIC 2020 test.
As for the notation FID/256, we denote KID/256 the KID computed over all 256x256 half overlapping patches in the evaluation dataset.

It is interesting to see that KID/256 paints a similar story to FID/256. MS-ILLM \cite{muckley2023improving} achieves state-of-the-art in KID/256 and remains unmatched by other methods. However, qualitatively we observe DIRAC-100 sample contains relevant details as seen in Section 5.1, Fig. 5.
While comparing LPIPS, we see DIRAC-1 and DIRAC-100 achieve similar LPIPS scores, further confirming its lower correlation with human judgement.

%%%%%%%%%%%%%%%%%%%%%%%%%%%%%%%%%%%%%%%%%%%%%%%%%%%%%%%%%%%%%%%
\subsection{Comparison to enhancement literature}
\label{appendix:additional_enhancement_comparison}

\begin{figure*}[t]
    \includegraphics[width=\textwidth]{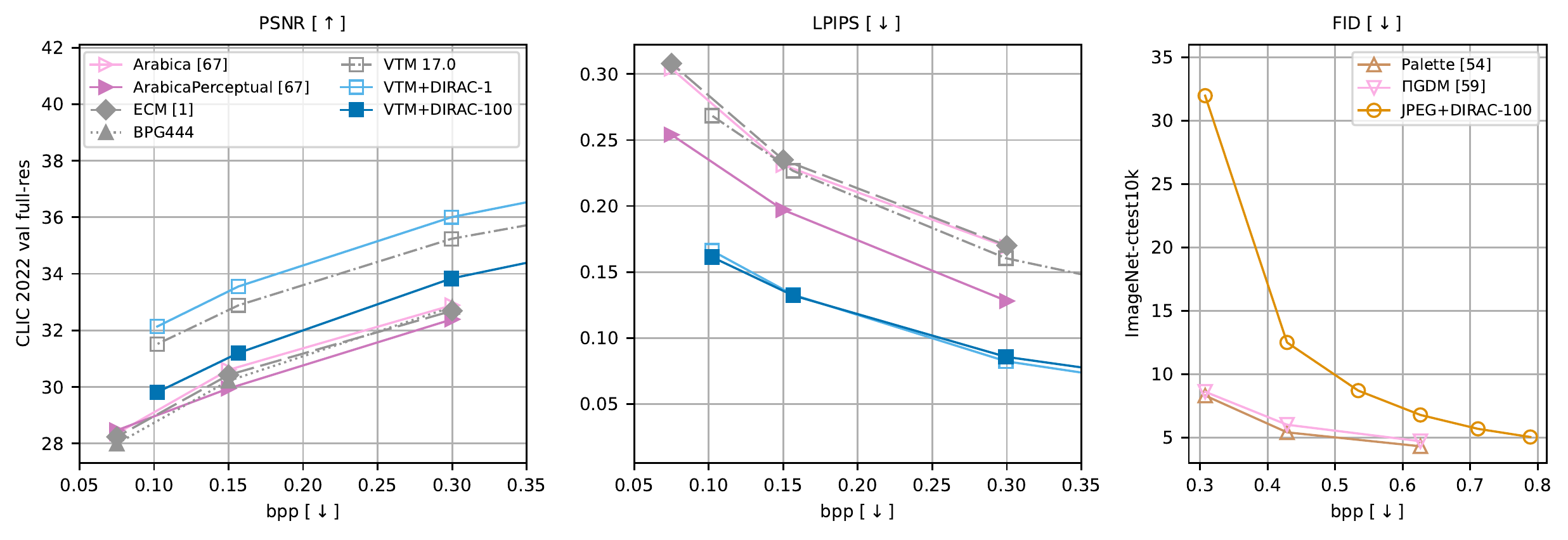}
    \caption{Additional quantitative comparisons of VTM+DIRAC and JPEG+DIRAC to enhancement literature on CLIC 2022 val and ImageNet-ctest10k respectively.
    Note that all models were trained on different datasets and have different number of parameters, which does not favor fair comparison. In particular Palette has about 5x the number of parameters as JPEG+DIRAC, was trained on ImageNet low-resolution dataset and is evaluated with 10x the number of sampling steps.
    }
    \label{fig:appendix:enhancement_additional_rp_curves}
\end{figure*}

In \cref{fig:appendix:enhancement_additional_rp_curves}, we report two additional sets of results comparing to enhancement literature on slightly different tasks and/or datasets.

The first two plots in \cref{fig:appendix:enhancement_additional_rp_curves} compare VTM+DIRAC to the latest standard codec Enhanced Compression Model (ECM) \cite{fraunhofer2022ecm} and its GAN-based learned in-loop filter methods (\ie enhancement) \emph{Arabic} and \emph{ArabicPerceptual} \cite{wang2022perceptual} on the CLIC 2022 val dataset. The work \emph{ArabicPerceptual} \cite{wang2022perceptual} is trained with a perceptual loss (LPIPS and discriminator), and is most relevant to our perceptual enhancement work \ie, to be compared to VTM+DIRAC-100.

The right-most plot in \cref{fig:appendix:enhancement_additional_rp_curves} compares JPEG+DIRAC to the diffusion-based image-to-image models \emph{Palette} \cite{saharia2022palette} and its follow-up work \emph{$\Pi$GDM} \cite{song2023solving}. Palette is explicitly trained for JPEG restoration, while $\Pi$GDM introduce an algorithm to re-purpose a pre-trained diffusion model to solve any inverse problem, including JPEG restoration, similar to \cite{kawar2022denoising,kawar2022jpeg}. While both outperform JPEG+DIRAC in terms of FID (not FID/256) on ImageNet-ctest10k, it is good to remember the comparison might not be the fairest. Both Palette and $\Pi$GDM were trained on ImageNet 256x256 resized crops, and hence cater well to low resolution datasets as well as being closer to the test data distribution. JPEG-DIRAC was trained on the much higher resolution CLIC 2020 train dataset. Furthermore, Palette has more than 5x the number of parameters of JPEG+DIRAC, and uses 10x the number of sampling steps.

%%%%%%%%%%%%%%%%%%%%%%%%%%%%%%%%%%%%%%%%%%%%%%%%%%%%%%%%%%%%%%%
\subsection{Noise schedule ablation}
\label{appendix:noise_schedule_ablation}

\begin{figure*}
    \centering
    \includegraphics[width=\textwidth]{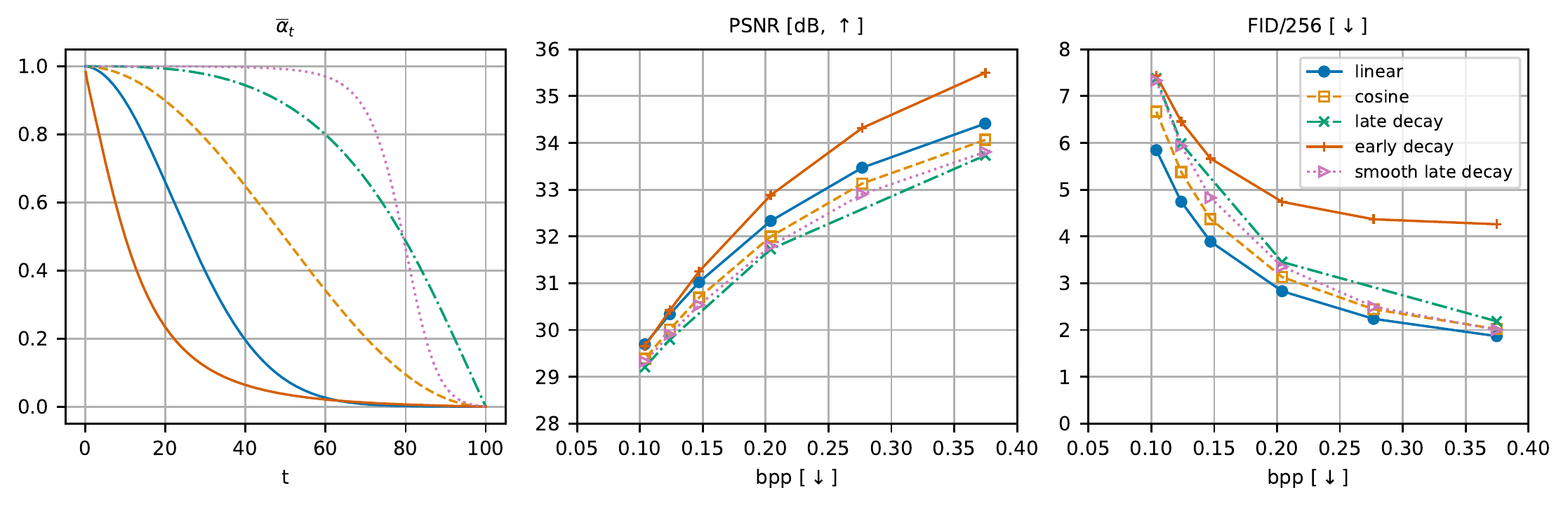}
   \caption{Shape of the $\overline{\alpha}_t$ curves for the different noise schedules (left panel) and their PSNR and FID/256 performance on CLIC 2020 test (center and right panel). Evaluation done with intermediate DIRAC checkpoints (150k steps).}
   \label{fig:appendix:noise_schedule_ablation}
\end{figure*}

The choice of noise schedule can be important for performance \cite{nichol2021improved}, and we tried several different options early in our experiments. The so-called \emph{linear} and \emph{cosine} schedules \cite{nichol2021improved} are two popular choices. Because some of our results suggested that our chosen linear schedule is not ideal (see discussion in \cref{appendix:ddim_sampling}), we also decided to try more extreme approaches, meaning schedules where the decay from sample to 0 (characterized by $\overline{\alpha}_t$) happens either early or very late. For those schedules we devised a generic formula that allows us to model a wide variety of shapes for $\overline{\alpha}_t$:

\begin{equation}
    \overline{\alpha}_t(L, p) = \frac{ \frac{1}{1+\exp(2L \cdot t^p - L)} - \overline{\alpha}_1(L, p) } {\overline{\alpha}_0(L, p) - \overline{\alpha}_1(L, p)}\;,
\end{equation}

where $L$ and $p$ are the parameters that define the shape of the decay, and a generalized time $t\in[0,1]$ is used. We define the following three schedules: 1) \emph{early decay} ($L=5,p=0.3$), which decays faster than the linear schedule; 2) \emph{late decay} ($L=1,p=3$), which decays later than the cosine schedule; and 3) \emph{smooth late decay} ($L=6,p=3$), which also decays late, but has close to zero gradient in the final forward steps. All schedules are visualized in \cref{fig:appendix:noise_schedule_ablation}, along with their performance in PSNR and FID/256. We find that the linear schedule offers the best tradeoff between fidelity and perceptual quality. Note that this evaluation was done with intermediate checkpoints (after 150k training steps), as we did not continue training with the other schedules.

%%%%%%%%%%%%%%%%%%%%%%%%%%%%%%%%%%%%%%%%%%%%%%%%%%%%%%%%%%%%%%%
\subsection{Compute}
\label{appendix:implementation:compute}

We provide information on the computational complexity of model components, and the cost of sampling, in Table \ref{tab:appendix:implementation:compute}.
We run benchmarking on a desktop workstation with a Nvidia 3080 Ti card, CUDA driver version is 455.32.00, CUDA 11.1.
We use 1,000 forward passes on square inputs of size $256\times256$ and $1024 \times 1024$, use GPU warmup, and record inference time using CUDA events. 
We do not include time to entropy code here.

\begin{table*}[h!]
    \centering
    \caption{Computational complexity. Runtime mean and standard deviation were obtained using 1,000 forward passes, either for a $256\times256$ or a $1024\times1024$ input. 
    Note that none of the models here were explicitly optimized for inference speed, numbers are indications only.}
    \label{tab:appendix:implementation:compute}
    \begin{tabular}{l r r r}
     \toprule
                           & Parameter count & Runtime 256 (ms) & Runtime 1024 (ms) \\
     \midrule
         SwinT-ChARM  &  28.4M & $37.9 \pm 0.5$ & $153.8 \pm 0.4 $ \\
         DIRAC U-Net       & 108.4M &  $37.5 \pm 0.8 $ & $ 486.3 \pm 1.2 $ \\
     \midrule
        HiFiC              & 181.5M & $17.0\pm0.4$ & $129.2 \pm 0.5$ \\
     \bottomrule
    \end{tabular}
\end{table*}

All our (base codec) multi-rate SwinT-ChARM models were trained on a single Nvidia V100 card.
Final DIRAC models were trained on 2 Nvidia A100 cards using data paralellism.

A fair comparison between methods would not only require equalization of bitrate, but ideally also equalization of training and test-time compute.
For example, HiFiC decoding complexity is substantially lower than that of DIRAC if many sampling steps are used, which will give DIRAC-100 an unfair advantage.  
DDPMs also see more datapoints than HiFiC during training as a larger batch size is typically used.
In this work, we mainly focused on feasibility of the approach, and have therefore not equalized training compute.

%%%%%%%%%%%%%%%%%%%%%%%%%%%%%%%%%%%%%%%%%%%%%%%%%%%%%%%%%%%%%%%
\subsection{Texture differences JPEG+DIRAC-1 vs. JPEG+DIRAC-100}
\label{appendix:texture_differences_jpeg}

\begin{figure*}
    \centering
    \includegraphics[width=0.85\textwidth]{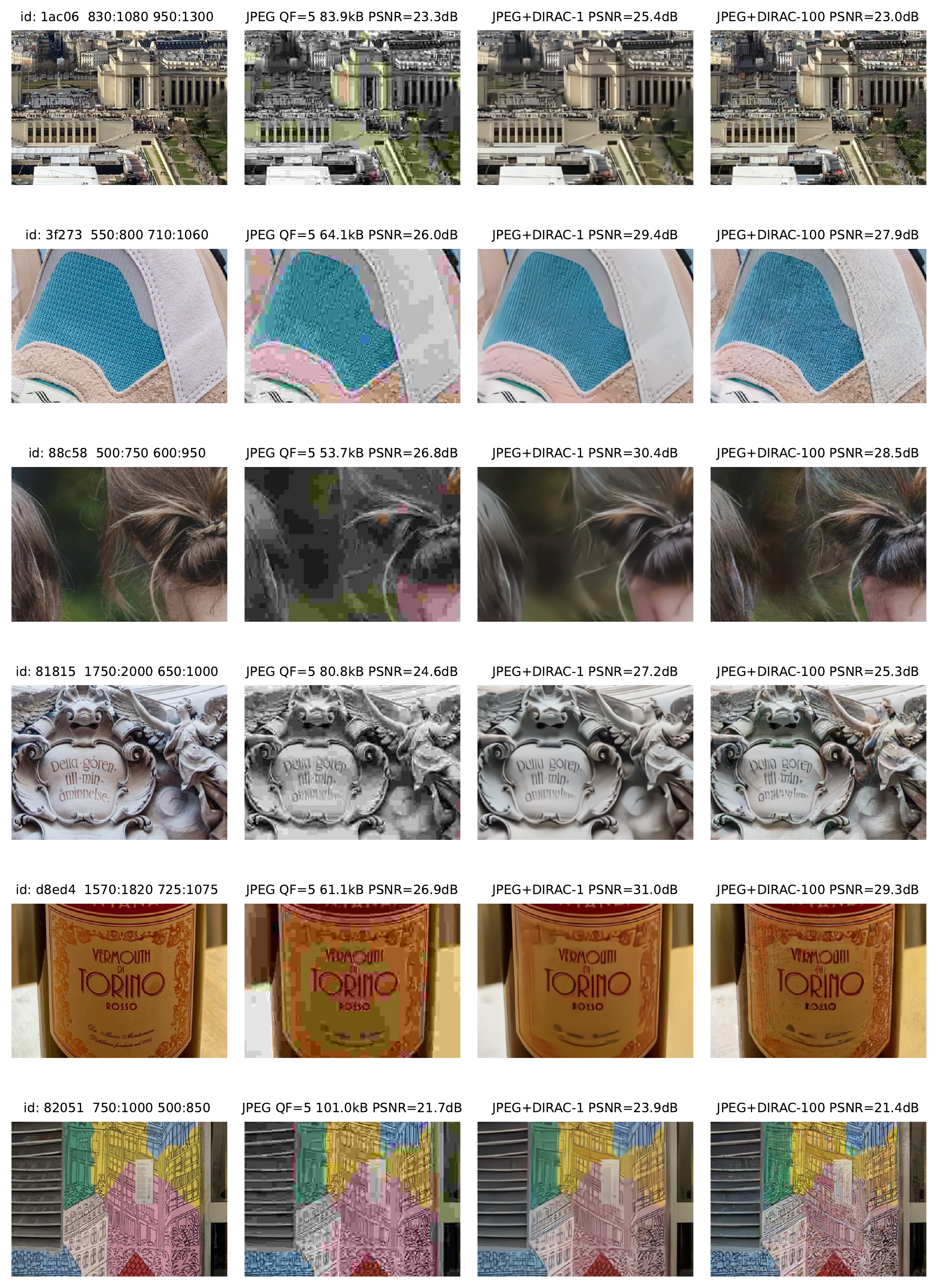}
    \caption{Samples from the JPEG enhancement model. Note the large texture differences between JPEG+DIRAC-1 and JPEG+DIRAC-100, and the overall drastic improvement over JPEG on CLIC test 2020 crops. Best viewed electronically.}
    \label{fig:appendix:texture_differences_jpeg}

\end{figure*}

In Sec. 5.2, in particular in Fig. 6, we observed little difference in LPIPS on ImageNet-val1k between JPEG+DIRAC-1 and JPEG+DIRAC-100 \ie, between 1 and 100 sampling steps. Upon qualitative assessment in \cref{fig:appendix:texture_differences_jpeg}, it becomes obvious how sampling drastically improves the textures and cleans JPEG artefacts. JPEG+DIRAC-1 gets rid of most of the blocking, and including removal / dampening of wrong color hues, like the typical purple artefact. One of the main failure modes of JPEG+DIRAC is small text and small faces---typical for generative compression models---and both can be seen in example ``81815''.

It is good to note that FID/256 on CLIC 2020 test does improve substantially, which better aligns with human judgement. It is yet another signal pointing towards FID/256 as the objective metric correlating best with human evaluation \cite{mentzer2020hific}. On the contrary, JPEG+DIRAC-100 always has worse PSNR than JPEG+DIRAC-1, as it measures fidelity but poorly correlates with human judgement.

\begin{figure*}
    \centering
    \includegraphics[width=0.85\textwidth]{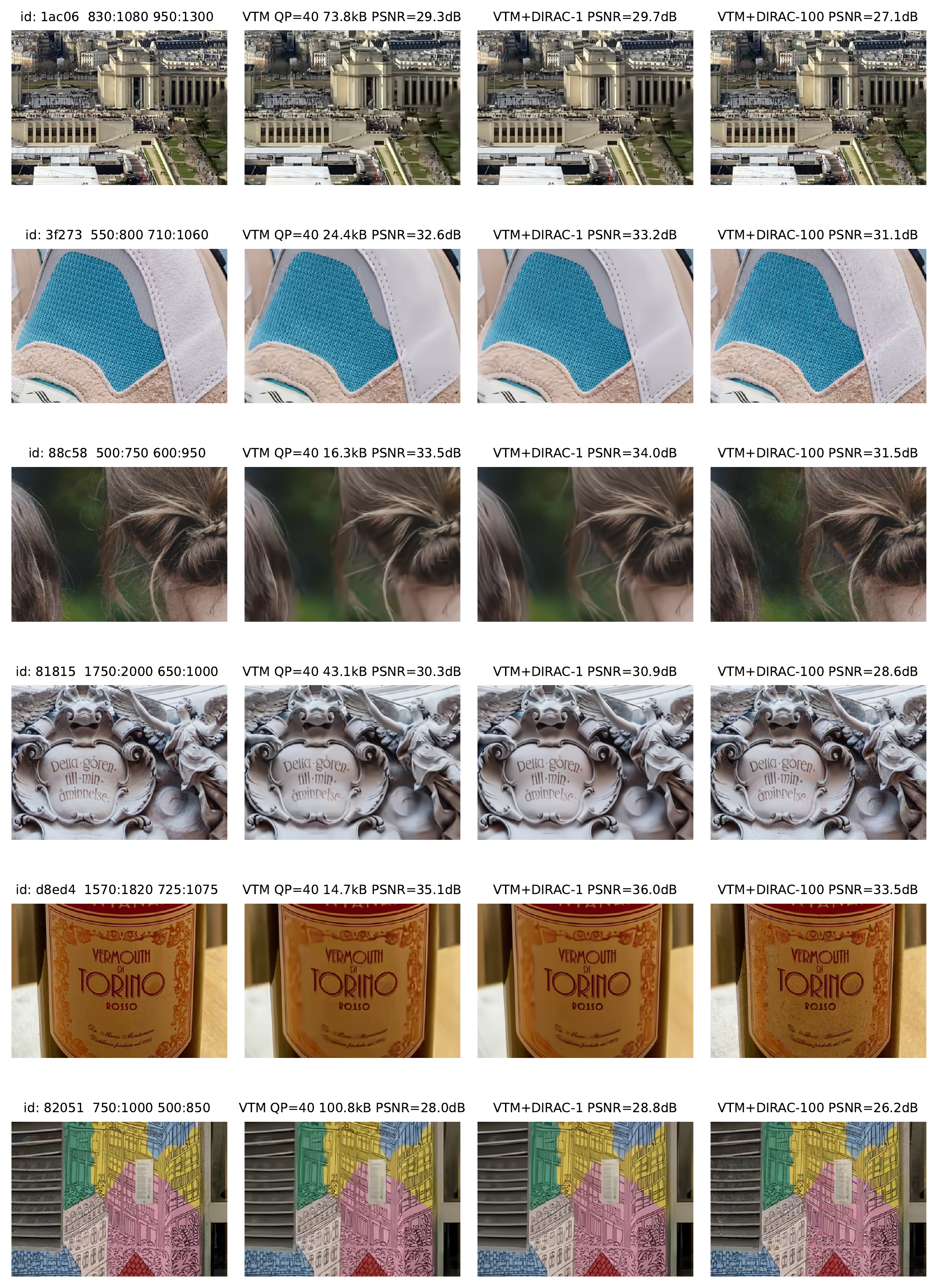}
    \caption{Samples from the VTM enhancement model. Note the large texture differences between VTM+DIRAC-1 and VTM+DIRAC-100, and the overall improvement over VTM on CLIC test 2020 crops. Best viewed electronically.}
    \label{fig:appendix:texture_differences_vtm}

\end{figure*}

Finally, for completeness we include the same set of samples for VTM+DIRAC in \cref{fig:appendix:texture_differences_vtm}. In most if not all cases human judgement would favor VTM+DIRAC-100, although it consistently has worse PSNR. We can see that, contrary to JPEG+DIRAC, small text is better handled. It is most likely due to much better conditioning / starting point. Yet some artefacts can still creep in, as in ``d8ed4'' where the wine label tends to be ``overtextured''.

%%%%%%%%%%%%%%%%%%%%%%%%%%%%%%%%%%%%%%%%%%%%%%%%%%%%%%%%%%%%%%%
\subsection{Qualitative results}
\label{appendix:qualitative_results}

\begin{figure*}
    \centering
    \begin{subfigure}{0.97\textwidth}
        \centering
        \includegraphics[width=1.0\textwidth,trim={0 5mm 0 0},clip]{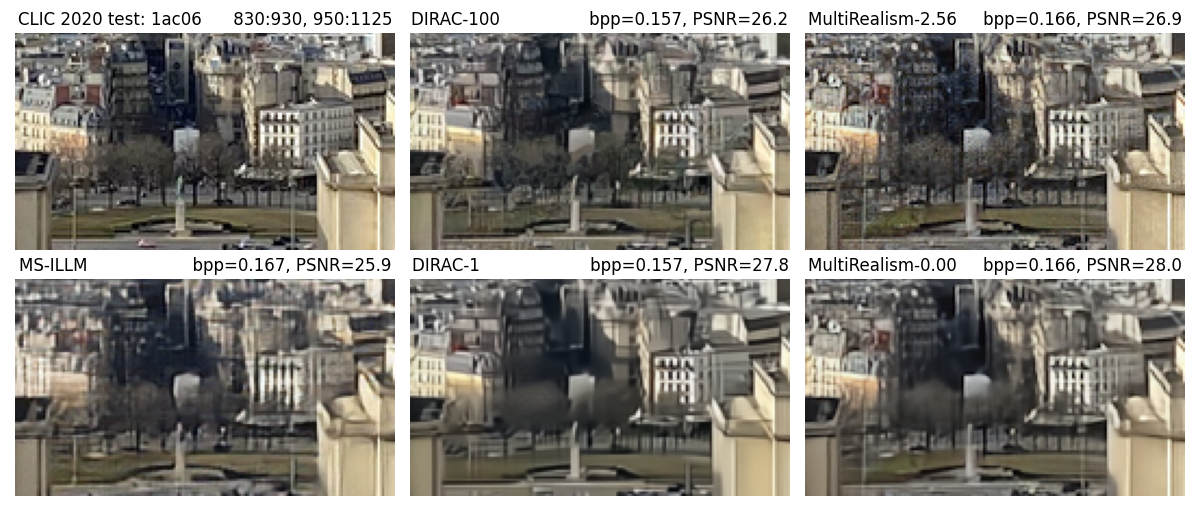}
    \end{subfigure}
    \begin{subfigure}{0.97\textwidth}
        \centering
        \includegraphics[width=1.0\textwidth,trim={0 5mm 0 0},clip]{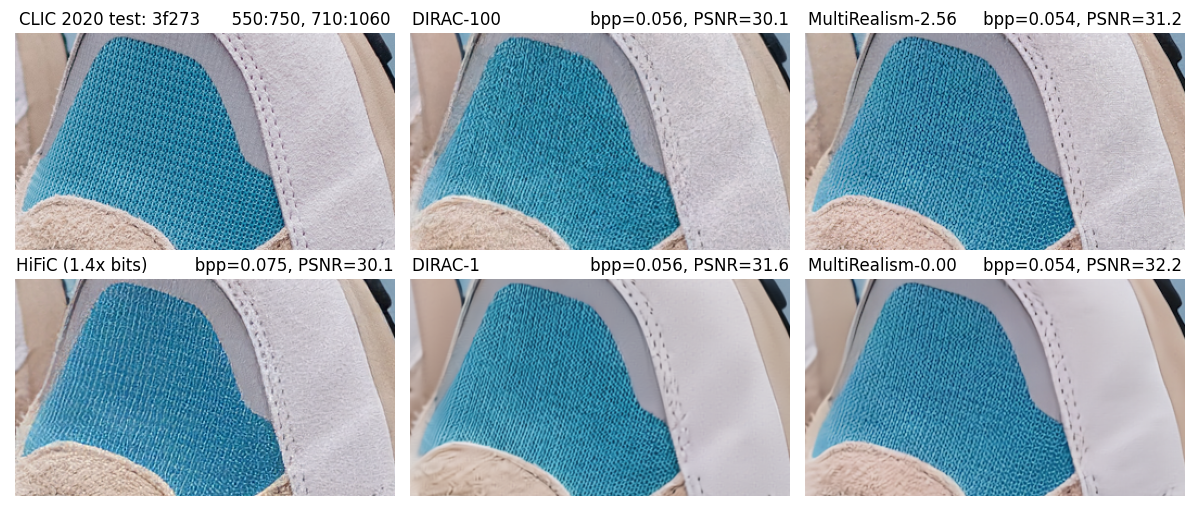}
    \end{subfigure}
    \begin{subfigure}{0.97\textwidth}
        \centering
        \includegraphics[width=1.0\textwidth,trim={0 5mm 0 0},clip]{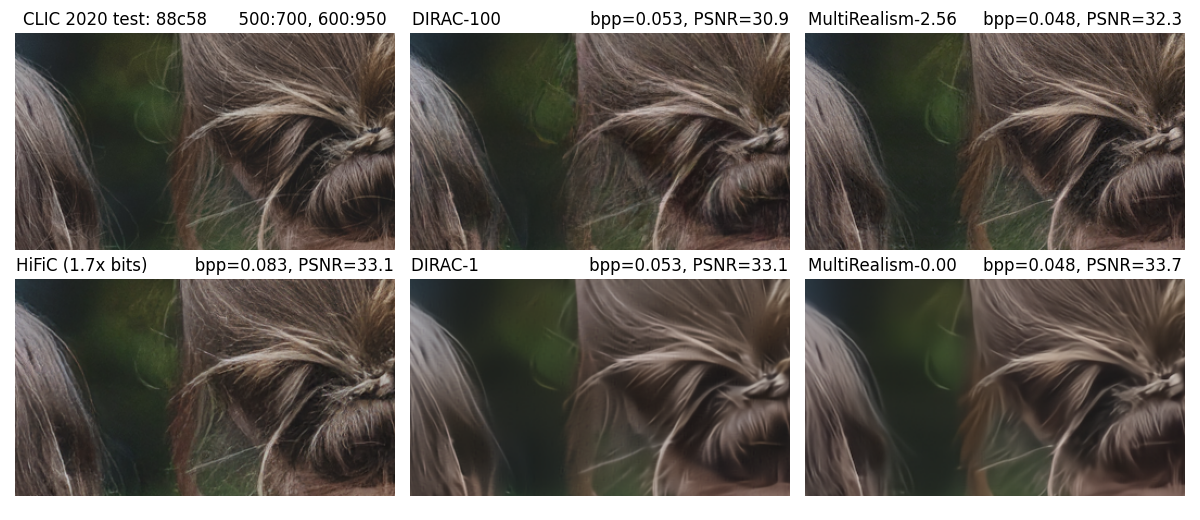}
    \end{subfigure}
    \caption{Reconstructions from various image codecs on CLIC 2020 test data. Crop locations were chosen to match related work to ensure a fair comparison \cite{agustsson2022multi, muckley2023improving}. Bitrates were chosen so that they match the Multirealism baseline \cite{agustsson2022multi}.}
    \label{fig:appendix:reconstructions1}
\end{figure*}

\begin{figure*}
    \centering
    \begin{subfigure}{0.97\textwidth}
        \centering
        \includegraphics[width=1.0\textwidth,trim={0 5mm 0 0},clip]{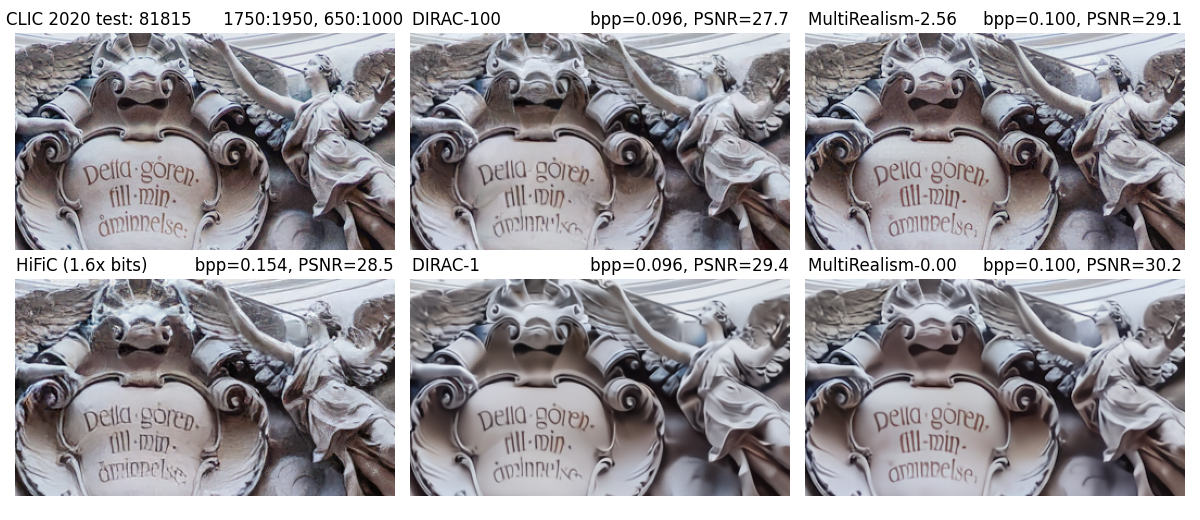}
    \end{subfigure}
    \begin{subfigure}{0.97\textwidth}
        \centering
        \includegraphics[width=1.0\textwidth,trim={0 5mm 0 0},clip]{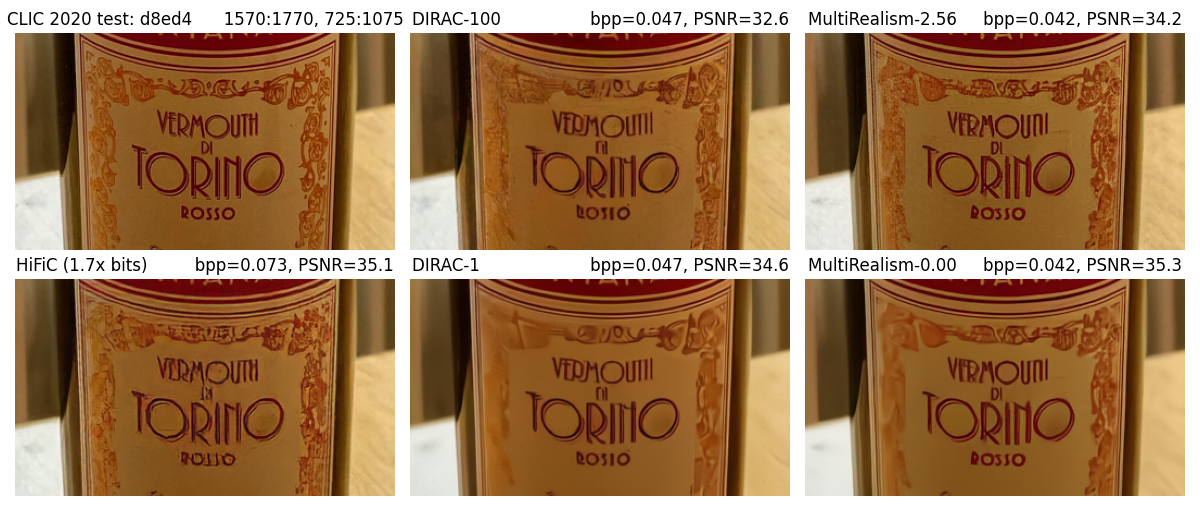}
    \end{subfigure}
    \begin{subfigure}{0.97\textwidth}
        \centering
        \includegraphics[width=1.0\textwidth,trim={0 5mm 0 0},clip]{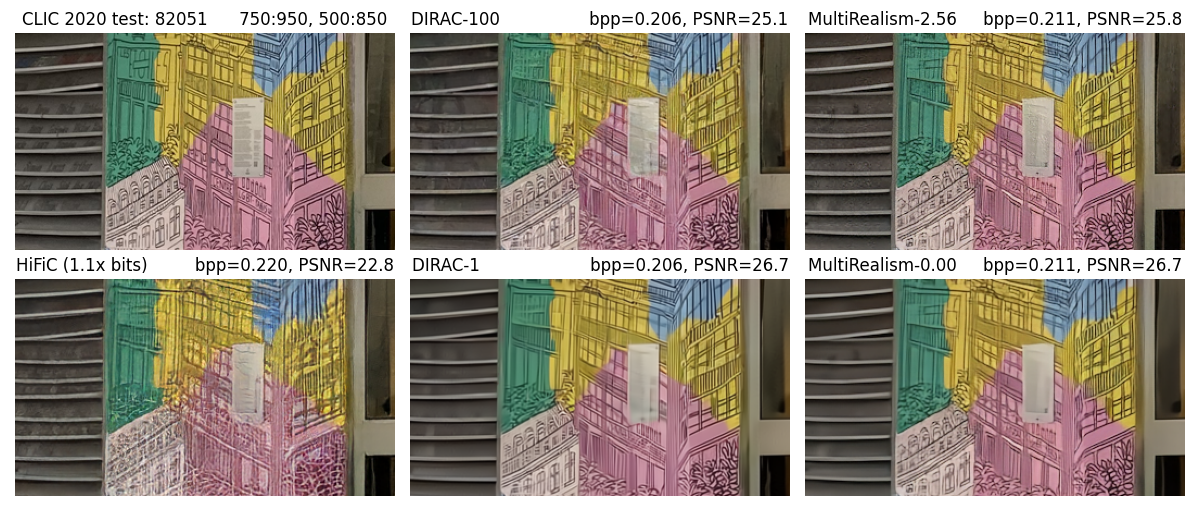}
    \end{subfigure}
    \caption{Reconstructions from various image codecs on CLIC 2020 test data. Crop locations were chosen to show text and salient features such as fine lines and small faces.}
    \label{fig:appendix:reconstructions2}
\end{figure*}

We include more qualitative results below. 
In \cref{fig:appendix:reconstructions1}, \cref{fig:appendix:reconstructions2}, we show crops from images from the CLIC 2020 test set and reconstructions by various image codecs. 
We include images chosen by \cite{agustsson2022multi, muckley2023improving} to ensure a fair comparison.
We match the bitrate to the Multirealism model of Agustsson \etal \cite{agustsson2022multi}.
The HiFiC reconstructions are obtained from a reimplemented HiFiC model, trained for the 0.15 bits per pixel range, and we do not perform bitrate matching.

\end{document}